\documentclass{article}

\usepackage{microtype}
\usepackage{graphicx}
\usepackage{subfigure}
\usepackage{booktabs} 

\usepackage{hyperref}

\usepackage[accepted]{icml2025}

\usepackage{amsmath}
\usepackage{amssymb}
\usepackage{mathtools}
\usepackage{amsthm}

\usepackage[capitalize,noabbrev]{cleveref}

\theoremstyle{plain}

\theoremstyle{definition}

\theoremstyle{remark}

\usepackage[textsize=tiny]{todonotes}
\usepackage{graphicx} 
\usepackage{float}  
\usepackage{fvextra}
\usepackage[most]{tcolorbox}

\usepackage{amsmath,amsfonts,bm}

\def\1{\bm{1}}

\RequirePackage{amsmath}
\RequirePackage{amssymb}
\RequirePackage{amsthm}
\RequirePackage{bm} 
\RequirePackage{url}
\usepackage{multirow}
\usepackage{natbib}
\usepackage{graphicx}
\usepackage{subfigure}
\usepackage{makecell}
\usepackage{booktabs}
\usepackage{array}
\usepackage{url}
\usepackage{algorithm}
\usepackage{algorithmic}
\usepackage{dsfont}

\usepackage{enumitem}

\usepackage{enumerate}
\usepackage[OT1]{fontenc}
\usepackage{natbib}

\usepackage{mathrsfs}

\usepackage{xcolor}
\usepackage{hyperref}

\def\##1\#{\begin{align}#1\end{align}}
\def\$#1\${\begin{align*}#1\end{align*}}

\let\tilde\widetilde

\newcommand{\cD}{\mathcal{D}}

\newcommand{\cM}{\mathcal{M}}

\usepackage{xcolor}

\definecolor{red1}{HTML}{f47983}
\definecolor{blue1}{HTML}{3eede7}
\definecolor{yellow1}{HTML}{f5dd6f}

\icmltitlerunning{BOLT}

\begin{document}

\twocolumn[
\icmltitle{BOLT: Bootstrap Long Chain-of-Thought in Language Models without Distillation}




\begin{icmlauthorlist}
\icmlauthor{Bo Pang}{yyy}
\icmlauthor{Hanze Dong}{yyy}
\icmlauthor{Jiacheng Xu}{yyy}
\icmlauthor{Silvio Savarese}{yyy}
\icmlauthor{Yingbo Zhou}{yyy}
\icmlauthor{Caiming Xiong}{yyy}
\\
Salesforce AI Research

\end{icmlauthorlist}

\icmlaffiliation{yyy}{Salesforce AI Research}

\icmlcorrespondingauthor{Bo Pang}{b.pang@salesforce.com}

\icmlkeywords{Machine Learning, ICML}

\vskip 0.3in
]



\printAffiliationsAndNotice{}  

\begin{abstract}
Large language models (LLMs), such as o1 from OpenAI, have demonstrated remarkable reasoning capabilities. o1 generates a long chain-of-thought (LongCoT) before answering a question. LongCoT allows LLMs to analyze problems, devise plans, reflect, and backtrack effectively. These actions empower LLM to solve complex problems. After the release of o1, many teams have attempted to replicate its LongCoT and reasoning capabilities. In terms of methods, they primarily rely on knowledge distillation with data from existing models with LongCoT capacities (e.g., OpenAI-o1, Qwen-QwQ, DeepSeek-R1-Preview), leaving significant uncertainties on systematically developing such reasoning abilities. In terms of data domains, these works focus narrowly on math while a few others include coding, limiting their generalizability. This paper introduces a novel approach to enable LLM’s LongCoT capacity without distillation from o1-like models or expensive human annotations, where we bootstrap LongCoT (BOLT) from a standard instruct model. BOLT involves three stages: 1) LongCoT data bootstrapping with in-context learning on a standard instruct model; 2) LongCoT supervised finetuning; 3) online training to further refine LongCoT capacities. In BOLT, only a few in-context examples need to be constructed during the bootstrapping stage; in our experiments, we created 10 examples, demonstrating the feasibility of this approach. We use Lllama-3.1-70B-Instruct to bootstrap LongCoT and apply our method to various model scales (7B, 8B, 70B). We achieve impressive performance on a variety of benchmarks, Arena-Hard, MT-Bench, WildBench, ZebraLogic, MATH500, which evaluate diverse task-solving and reasoning capabilities.
\end{abstract}

\section{Introduction}
\label{introduction}

\begin{figure*}[!t]
    \centering
    \includegraphics[width=\textwidth]{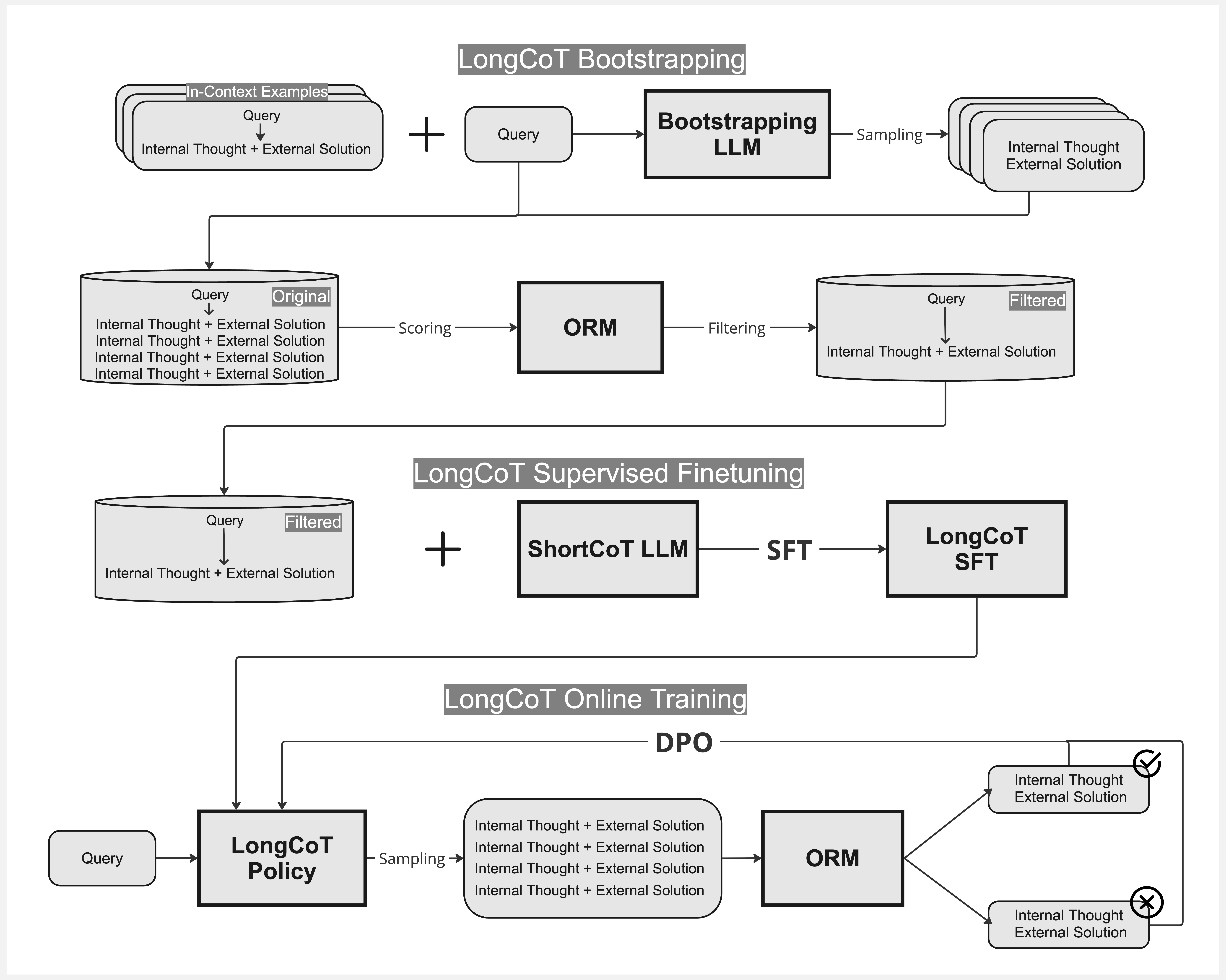} 
    \caption{Illustration of bootstrapping long chain-of-thought in large language models (BOLT). BOLT comprises three stages: 1) \textbf{LongCoT Bootstrapping} which involves synthesizing LongCoT data, 2) \textbf{LongCoT Supervised Finetuning} where we train a ShortCoT model to adapt to the LongCoT format, incorporating reasoning elements and practicing extended chains of thought before arriving at an external solution, 3) \textbf{LongCoT Online Training} where the LongCoT SFT model is further improved through online exploration and refinement. \textbf{Bootstrapping LLM} is a ShortCoT LLM that is used to generate LongCoT data via in-context learning. \textbf{ORM} is an outcome reward model which scores the external solution in the model response.}
    \label{fig:bolt}
\end{figure*}

Large language models (LLMs), such as OpenAI's o1 model, have exhibited extraordinary reasoning abilities, particularly on coding and mathematical problems. The o1 model employs long chain-of-thought (LongCoT), which involves generating an extended reasoning sequence prior to delivering a final answer. 
LongCoT enables LLMs to analyze problems, make plans, branch to different approaches, evaluate their reasoning through reflection, and, when necessary, backtrack to correct errors. By leveraging these problem-solving techniques via long chain-of-thought, LLMs can tackle highly intricate and multifaceted challenges, showcasing their potential to function as powerful tools for complex reasoning tasks across various disciplines.

In fact, almost all modern LLMs are able to reason through chain-of-thought via prompting techniques~\cite{wei2022chain}. Recent instruct models are trained on chain-of-thought data, making this chain-of-thought process their default mode, particularly when solving math problems~\cite{jiang2023mistral7b, grattafiori2024llama3herdmodels, gemmateam2024gemma2improvingopen}.
Additionally, \citet{wang2024chainofthoughtreasoningprompting} show that chain-of-thought is inherent in pre-trained LLMs. However, regular LLMs exhibit shorter and, more critically, simpler behavior in chain-of-thought, compared to o1-like models. In this paper, we refer to o1-like models that generate long chain-of-thought with rich reasoning behavior as \textit{LongCoT} models, while regular LLMs are referred to as \textit{ShortCoT} models.

Following the release of o1, numerous research teams have sought to replicate its LongCoT and reasoning capabilities. Methodologically, these efforts primarily rely on knowledge distillation using data derived from existing LongCoT models (e.g., OpenAI-o1, Qwen-QwQ, DeepSeek-R1-Preview). However, this approach leaves significant gaps
in understanding how to systematically develop such LongCoT reasoning skills. 
While distillation provides a shortcut for training LongCoT models, developing such models without relying on existing LongCoT models remains a black-box.
Regarding data domains, most of these studies concentrate on mathematical problem solving, with only a few expanding into coding tasks, limiting the broader applicability and generalization of their findings.

This paper introduces a novel approach to enable LLMs to develop LongCoT capabilities without relying on distillation from LongCoT models or expensive human annotations. Our method, called Bootstrapping LongCoT (BOLT), builds these capacities from a ShortCoT LLM through three key stages: (1) LongCoT data bootstrapping using in-context learning with a ShortCoT LLM, (2) LongCoT supervised finetuning, and (3) online training to further refine LongCoT skills. In the bootstrapping stage, only a minimal number of in-context examples are required—in our experiments, we created just 10 examples—highlighting the feasibility and efficiency of this approach. Using Llama-3.1-70B-Instruct as the bootstrapping model, we applied BOLT to various model scales (7B, 8B, 70B), achieving remarkable performance across diverse benchmarks, including Arena-Hard, MT-Bench, WildBench, ZebraLoigc, and MATH500.
These benchmarks cover: 1) challenging real-user queries involving information-seeking, creative writing, coding, planning, and math, 2) classical logic puzzles, and 3) competition-level math problems. They test a broad spectrum of reasoning and task-solving skills, demonstrating BOLT's effectiveness in enhancing LongCoT capabilities.

Unlike black-box distillation, BOLT represents a white-box approach to developing LongCoT reasoning in LLMs. To support future research, we will open-source our training data and recipes and trained models. In summary, our work provides a principled, cost-effective pathway for cultivating LongCoT reasoning skills from ShortCoT models.

\section{Related Work}

\paragraph{LongCoT and o1-like Model} 
OpenAI's o1 model \citep{jaech2024openai} employs long chain-of-thoughts, allowing it to leverage rich reasoning actions, such as branching, reflection, verification, to tackle complex problems before arriving at a final answer \citep{dutta2024think}. This approach enhances the model performance in areas such as mathematics, coding, and scientific problems. 
The LongCoT approach aligns with System 2 cognition \citep{kahneman2011thinking}, a mode of deliberate and sequential reasoning that mirrors human problem-solving strategies. By integrating reinforcement learning, o1 can refine its reasoning process dynamically, evaluating multiple solution paths, backtracking when necessary, and improving its approach through iterative self-correction. The shift towards deliberative reasoning represents an important trend in AI research, aiming to make LLMs more transparent, interpretable, and adaptable in complex decision-making scenarios \citep{ackoff1994systems, kahneman2011thinking}.

Despite o1's success, most existing attempts to replicate LongCoT rely on knowledge distillation and manually curated datasets \citep{min2024imitate, huang2024o1}. These approaches pose several challenges: they often fail to generalize beyond the specific training data, require access to high-quality reference models, and lack principled methods for directly training LongCoT reasoning from scratch. A concurrent work by DeepSeek \citep{guo2025deepseek} demonstrated that reinforcement learning applied to a 671B-parameter model can yield LongCoT capabilities. However, such large-scale models introduce significant computational barriers, making broad adoption and reproducibility infeasible. Furthermore, while DeepSeek provides significant transparency regarding their approach, some crucial details, particularly their data curation strategies, remain unclear.

\paragraph{LLM Reinforcement Learning and Self-Improvement}
Reinforcement learning has become a core approach for enhancing LLMs during post-training, particularly for improving the quality of model outputs. Traditional RL algorithms like Proximal Policy Optimization (PPO) \citep{schulman2017ppo} have been effective but computationally expensive, making them less feasible in resource-limited settings. 
Recent efforts have proposed some efficient alternatives. 

Rejection sampling techniques \citep{zelikman2022star,dong2023raft,gulcehre2023reinforced} filter and select the best responses from multiple model-generated candidates, improving training efficiency without requiring full policy optimization. Similarly, Direct Preference Optimization (DPO) methods \citep{rafailov2023direct,munos2023nash,ethayarajh2024kto} bypass explicit reward modeling by directly optimizing on preference data, achieving performance comparable to PPO at significantly lower training costs. Meanwhile, REINFORCE-based approaches \citep{ahmadian2024back, li2023remax,ahmadian2024back,shao2024deepseekmath} further streamline training by eliminating the value function, reducing memory and computation requirements.
These approaches have proven useful in downstream tasks, yielding measurable gains in accuracy and coherence.
Building on them, recent work has explored self-improving LLMs, where models iteratively refine their own outputs using generated feedback loops \citep{xu2023some,snorkelai@pair,xiong2023iterative,yuan2024self,dong2024rlhf,guo2024direct}. These approaches enable LLMs to autonomously evaluate, critique, and improve their responses over multiple iterations, integrating self-feedback into the training process. This fosters an adaptive learning cycle, allowing models to progressively enhance reasoning depth, factual accuracy, and coherence over time.

Despite these advances, most existing RL-based methods focus on single-stage response generation, where models directly produce final answers without refining intermediate reasoning steps (e.g., internal states or “thoughts”). While inference-time techniques like chain-of-thought prompting \citep{wei2022chain} and self-consistency decoding \citep{wang2022self} have shown that explicit intermediate reasoning improves accuracy, such multi-stage reasoning is rarely incorporated into training itself. Recent work on deliberation-based methods \citep{madaan2023selfrefineiterativerefinementselffeedback} suggest that iterative refinement enhances reasoning quality, but most RL-based approaches lack mechanisms for models to revise, backtrack, or critique their own internal thought processes. As a result, current models struggle to recover from early reasoning errors or refine suboptimal strategies, limiting their robustness and adaptability.

\section{BOLT}
This paper introduces \textbf{BOLT} for learning LongCoT models by bootstrapping long-form chain-of-thought from ShortCoT models. BOLT comprises of three stages: 1) LongCoT Bootstrapping which involves synthesizing LongCoT data, 2) LongCoT Supervised Finetuning where we train a ShortCoT model to adapt to the LongCoT format, incorporating reasoning elements and practicing extended chains of thought before arriving at an external solution, 3) LongCoT online training where the LongCoT SFT model is further improved through online exploration and onpolicy refinement. An overview of BOLT is depicted in Figure~\ref{fig:bolt}.

Before discussing the method in detail, we introduce key notations. Let $x$ represent a query, $z$ denote internal thoughts, and $y$ indicate an external solution. Additionally, we use $\cM$ to denote off-the-shelf LLMs and $\pi$ for models or policies trained in our experiments.

\subsection{LongCoT Bootstrapping}
\label{sec:longcot_bootstrapping}
In our earlier experiments, we investigated various approaches for constructing LongCoT data such as prompt engineering on ShortCoT models and employing multi-agent systems (with actor agent and judge agent). However, these approaches were neither stable nor reliable. To address this, we developed a simple yet effective method for generating LongCoT data by bootstrapping ShortCoT LLMs.

\begin{figure}[t]
    \centering
    \includegraphics[width=\columnwidth]{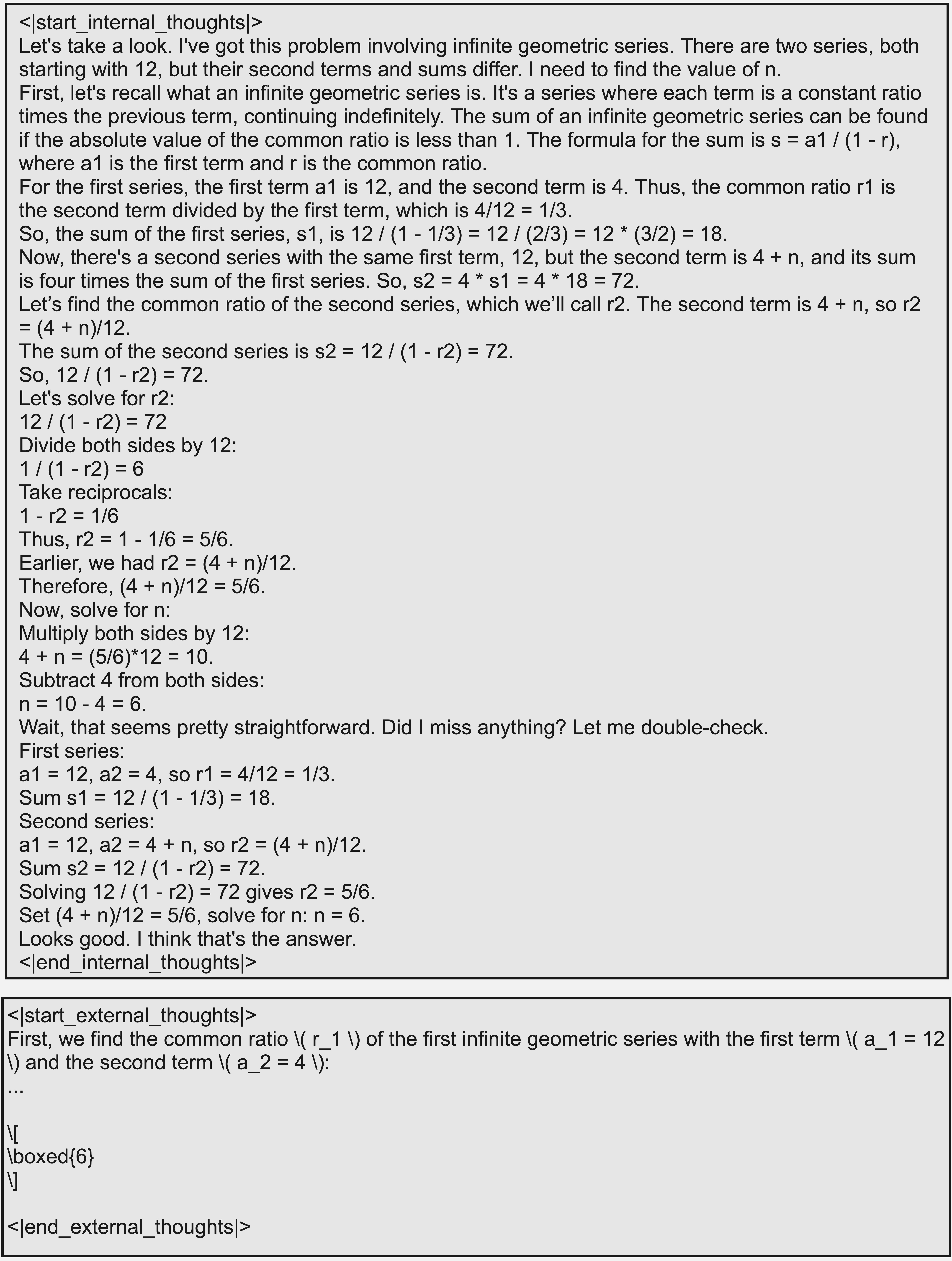}
    \caption{An illustration of long chain-of-thought as internal thoughts. Portions of the external solution are omitted for brevity.}
    \label{fig:output_example}
\end{figure}

\subsubsection{LongCoT with In-Context Learning}
ShortCoT models demonstrate some capability to produce chain-of-thought reasoning and handle complex tasks. To induce LongCoT, we leverage in-context examples of LongCoT to prompt ShortCoT models. These in-context examples guide the models to generate long-form chain-of-thought reasoning. Our findings reveal that as long as the ShortCoT language model is sufficiently strong, the generated responses are of reasonable quality and serve as a good basis for further processing.

To construct the in-context examples, each instance includes a long-form chain-of-thought and its corresponding solution derived from the reasoning process. See Figure~\ref{fig:output_example} for an example. The LongCoT incorporates essential reasoning actions: problem analysis, planning, branching, and reflection. These actions mirror key elements of human reasoning. By illustrating these actions via the in-context examples, the in-context learning capacity of ShortCoT models enable them to emulate LongCoT reasoning processes effectively. We collect $10$ in-context learning examples where each example consists of a query ($x$), a chain of internal thoughts ($z$), and an external solution ($y$). Let's denote the collection of in-context examples as $\cD_{\text{ICL}} = \{(x, y, z)\}$.

\begin{figure}[t]
    \centering
    \includegraphics[width=\columnwidth]{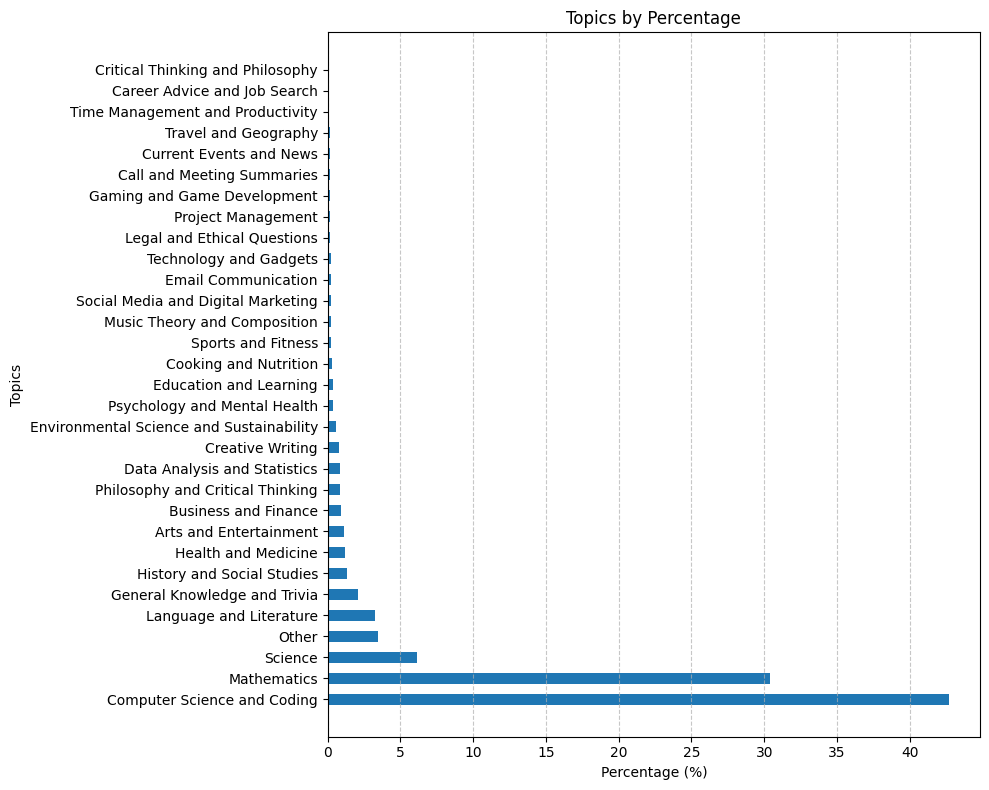} 
    \caption{Topic distribution of query data in LongCoT Bootstrapping, $\cD_{\text{b-query}}$.}
    \label{fig:topic_dist}
\end{figure}

\subsubsection{Query Mixture Curation}
\label{sec:query_curation}

While most prior works focus primarily on math problem solving, we believe that reasoning would benefit generic tasks and improve an LLM's helpfulness in general. Thus, we construct a query distribution that covers a wide range of topics and shift the distribution towards harder queries. Our query curation pipeline involves three steps: 1) query collection, 2) difficulty scoring and filtering, and 3) topic tagging and sub-sampling.

We first collect a large set of high-quality instruction datasets from public sources, such as ShareGPT \citep{vicuna2023}, SlimOrca \citep{SlimOrca}, MathInstruct \citep{yue2023mammoth}, and Evol-Instruct \citep{xu2023wizardlm} (see the Appendix for a full list). Only the query (of the first turn if it is a multi-turn chat) of each data instance is retained.
We next assign a difficulty level to each query. We follow the approach introduced by LMSys Team~\cite{arenahard2024} to select high quality user queries where seven criteria are considered: specificity, domain knowledge, complexity, problem-solving, creativity, technical accuracy, and real-world application. We assign a binary (0/1) label to each query on each criterion, and the quality or difficulty level of each query is determined by the total over the seven criteria. We keep queries with a score greater than or equal to $5$. 
Third, we use a pipeline to assign a topic to each query.
We identify a list of high-level topics by analyzing a subset of queries by LLM and human annotator. Then an LLM is employed to assign each query a topic from the list. We subsample the dataset based on the topic distribution. Figure~\ref{fig:topic_dist} displays the topic distribution after subsampling. Note that coding and math problems still dominate the query mixture after subsampling. This is due to two reasons: 1) coding and math problems are generally harder and 2) coding and math problem dominates our data sources, public instruct data, due to current research community's interest and their relative ease of data curation. We denote the set of queries from this step as $\cD_{\text{b-query}} = \{x\}$ where b-query indicates bootstrapping query.

\begin{figure}[t]
    \centering
    \includegraphics[width=\columnwidth]{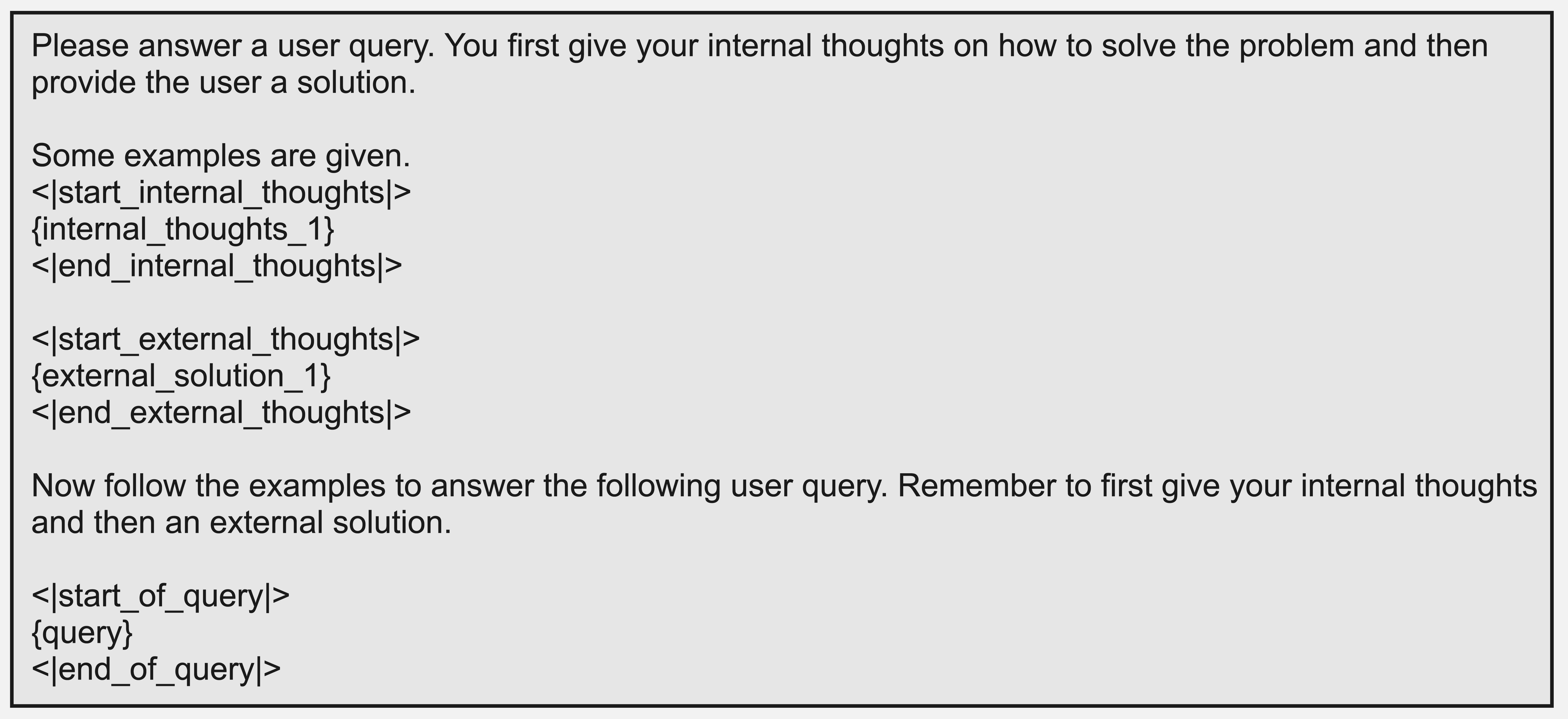}
    \caption{An illustration of the prompt used in LongCoT Bootstrapping.}
    \label{fig:icl_instructions}
\end{figure}

\subsubsection{Response Generation}
Given in-context examples, $\cD_{\text{ICL}}$, and the query set, $\cD_{\text{b-query}}$, we sample $n$ responses ($y, z$) from $\cM_{\text{bootstrapping}}$, in particular,
\begin{equation}
    (y, z) \sim \cM_{\text{bootstrapping}}(y, z | f_\text{formatting}(x, \tilde{\cD}_{\text{ICL}})),
\end{equation}
where $x \sim \cD_{\text{b-query}}$, $\tilde{\cD}_{\text{ICL}}$ is a subset of $\cD_{\text{ICL}}$, $f_\text{formatting}$ is a template that wraps $x$ and $\tilde{\cD}_{\text{ICL}}$ as an LLM input (see Figure~\ref{fig:icl_instructions}). In our experiments, $n=8$, $\lvert \tilde{\cD}_{\text{ICL}} \rvert = 3$, and $\cM_{\text{bootstrapping}}$ is Llama-3.1-70B-Instruct~\cite{grattafiori2024llama3herdmodels}. $\cM_{\text{bootstrapping}}$ is a ShortCoT LLM, but its basic reasoning capacity and generic instruction-following capacity enable it to generate responses following the format of long chain-of-thoughts and demonstrating reasoning elements in the thoughts. The procedure indeed requires a strong enough (in terms of reasoning and instruction-following capacity)  $\cM_{\text{bootstrapping}}$. Empirically, while Llama-3.1-8B-Instruct cannot reliably generate LongCoT responses following instructions and examples, Llama-3.1-70B-Instruct work well. Let's denote the sampled responses together with the queries as $\cD_{\text{bootstrapping}}^{(\text{original})} = \{(x, \{y_i, z_i\}_{i=1}^n\})$

\subsubsection{Response Filtering}
\label{sec:data_filtering}
While $\cD_{\text{bootstrapping}}^{(\text{original})}$ are of reasonable quality, we conduct filtering steps to further improve the LongCoT data. We first use some heuristics and rules to filter out data where the responses $(y, z)$ does not follow the particular format as demonstrated in the in-context examples (see Figure~\ref{fig:output_example}).  

We next filter out data with low-quality responses. Each response consists of $z$ (internal thoughts) and $y$ (an external solution). 
We don't have access to a judge or reward model on $z$ (while training such a reward model would require LongCoT data in the first place). However, many reward models and related data on judging the quality of $y$ have been published. These models can be viewed as outcome reward models (ORM) for a response with $(z, y)$. Therefore we use ORM to access the quality of $y$ and filtering data instance based on its quality score.
With an ORM, we have a quality distribution of all $y's$ from $\cD_{\text{bootstrapping}}^{(\text{original})}$. We first remove all data instance with score lower than 30th percentile of the score distribution and then choose the response with highest $y$ score. After the filtering steps, we obtain a high quality LongCoT data $\cD_{\text{bootstrapping}} = \{(x, y, z\})$.

\subsection{LongCoT Supervised Finetuning}
With $\cD_{\text{bootstrapping}} = \{(x, y, z\})$, we can conduct supervised finetuning on a ShortCoT model to allow it learn long form chain-of-thought and reasoning elements involved in it and the format of first producing internal thoughts and then an external response. Note the ShortCoT model does not necessarily need to be $\cM_{\text{bootstrapping}}$ but can be other models too. In our experiments, we apply LongCoT Supervised Finetuning with $\cD_{\text{bootstrapping}}$ to various models. Supervised finetuning leads to an initial LongCoT model, $\pi_{0}$.

\subsection{LongCoT Online Training}
With the SFT model $\pi_{0}$ as an initialization, we conduct online training to further improve the policy, $\pi_\theta(y, z \mid x)$, and it involves,

\begin{align}\label{eq:rl}
\max_{\pi_{\theta}}  &\mathbb{E}_{x\sim \cD_{\text{online}}, y,z\sim \pi_{\theta}(y, z \mid x)} \bigl[ \nonumber
r_{\phi}(x, z, y)\bigr] - \\
&\beta\mathbb{D}_{\textrm{KL}}\bigl[\pi_{\theta}(y, z\mid x)\mid \mid \pi_{0}(y, z\mid x)\bigr],
\end{align}
where $r_{\phi}(x, z, y)$ is a reward model. Similar to the strategy used in Section~\ref{sec:data_filtering} Response Filtering, we use an outcome reward model, which assign a score to $y$ given $x$, that is, $r_\phi: \mathcal{X} \times \mathcal{Y} \rightarrow \mathbb{R}$. In practice, we also include a rule-based format reward to facilitate model response following the defined format (see Figure~\ref{fig:output_example}).


The generic reward maximization with conservative constraint objective (Equation~\ref{eq:rl}) can be instantiated with several variants such as DPO, REINFORCE, RLOO, PPO. In our experiments, DPO works the best in terms of performance and efficiency and we choose DPO for our model training. See Section~\ref{sec:ablation_algorithm} for an ablation.





\section{Experiments}
In this section, we demonstrate that BOLT is a highly effective approach to develop LongCoT capacities in ShortCoT LLMs. We begin with a comprehensive evaluation of BOLT across diverse benchmarks and multiple model scales,
followed by a series of ablation studies to provide deeper insights into the effectiveness of our approach.

\subsection{Experiment Setup}
\paragraph{Evaluation Benchmarks} 

We focus on evaluating models’ reasoning capabilities across diverse domains, with an emphasis on real-world queries. MT-Bench~\cite{zheng2023judging} is a widely used benchmark that covers multi-turn questions spanning eight domains: writing, roleplay, reasoning, math, coding, information extraction, STEM, and the humanities. Arena-Hard~\cite{li2024crowdsourced} comprises challenging prompts drawn from crowd-sourced datasets featuring real user queries, including ChatBot Arena~\cite{li2024crowdsourced} and Wildchat-1M~\cite{zhao2024wildchat1mchatgptinteraction}. A significant portion of Arena-Hard prompts involves real-world coding problems. To minimize the influence of response length and markdown formatting, we use the style-controlled version of Arena-Hard, known as Arena-Hard-SC, as our focus is on the substance of model responses rather than their writing style. WildBench~\cite{lin2024wildbench} further complements this evaluation by including challenging real-world queries selected from over one million human-chatbot conversation logs. ZebraLogic~\cite{zebralogic2024} specifically targets logical reasoning, with each example being a logic grid puzzle, a typical Constraint Satisfaction Problem (CSP) commonly used to assess human reasoning in exams like the LSAT. Lastly, MATH500~\cite{lightman2023letsverifystepstep} is a representative subset of the MATH dataset~\cite{hendrycks2021measuring}, featuring problems drawn from various mathematics competitions, including the AMC and AIME.

\paragraph{Models}
We apply BOLT to three models, Mistral-7B-Instruct-v0.3~\cite{jiang2023mistral7b}, Meta-Llama-3.1-8B-Instruct~\cite{grattafiori2024llama3herdmodels}, Meta-Llama-3.1-70B-Instruct~\cite{grattafiori2024llama3herdmodels}, to test the effectiveness of our method across different model scales. Besides instruct models, we also tested a base model, Meta-Llama-3.1-8B-base, as the initial model for our method and observe similar enhancing effects. See Section~\ref{sec:ablation_init} for an ablation study.

\paragraph{Training Hyperparameters}
In the first stage of BOLT, LongCoT Bootstrapping (see Figure~\ref{fig:bolt}) generates a dataset of 220k instances. LongCoT supervised finetuning is performed on this dataset for 4 epochs, with the final checkpoint used for the next training stage. Hyperparameters (same for all models) include a maximum sequence length of 16,384, a batch size of 128, a learning rate of 2e-5, a cosine learning rate scheduler with a warm-up ratio of 0.1, and AdamW~\cite{loshchilov2019decoupledweightdecayregularization, kingma2014adam} as the optimizer. The training is conducted using Axolotl~\cite{axolotl}. Mistral-7B and Llama-8B are trained on a single 8xH100 node, requiring about 6 hours, while Llama-70B is trained on eight 8xH100 nodes, completing in about 5 hours.

The LongCoT online training involves sampling from a policy model, where sampling temperature is $1.0$ and top-p is $1.0$. Eight samples are sampled given each query. To assign a reward to each online sample (external solution in our method), we use ArmoRM-Llama3-8B~\cite{ArmoRM} as the reward model. An ablation study on the choice of reward model is presented in Section~\ref{sec:ablation_reward}.
DPO trianing hyperparameters include a regularization coefficient of $\beta = 0.1$, a learning rate of 5e-7 with a cosine scheduler and a warm-up ratio of 0.1, a batch size of 128, and AdamW as the optimizer.
Online training is conducted over $3$ iterations and each iteration consists of $2$ epochs.
Each iteration uses 33k hard prompts selected from a list of open-sourced preference data (see Appendix for the full list). The queries used in our experiments will be open-sourced as well.
Our DPO training is based on an open-sourced library TRL~\cite{vonwerra2022trl}.
Mistral-7B and Llama-8B are trained on one 8xH100 node for about $14$ hours. Llama-70B is trained on eight 8xH100 for about $20$ hours.


\subsection{Main Results}
\label{sec:main}

\begin{figure}[H]  
    \centering
    \subfigure[Mistral-7B]{
        \includegraphics[width=\columnwidth]{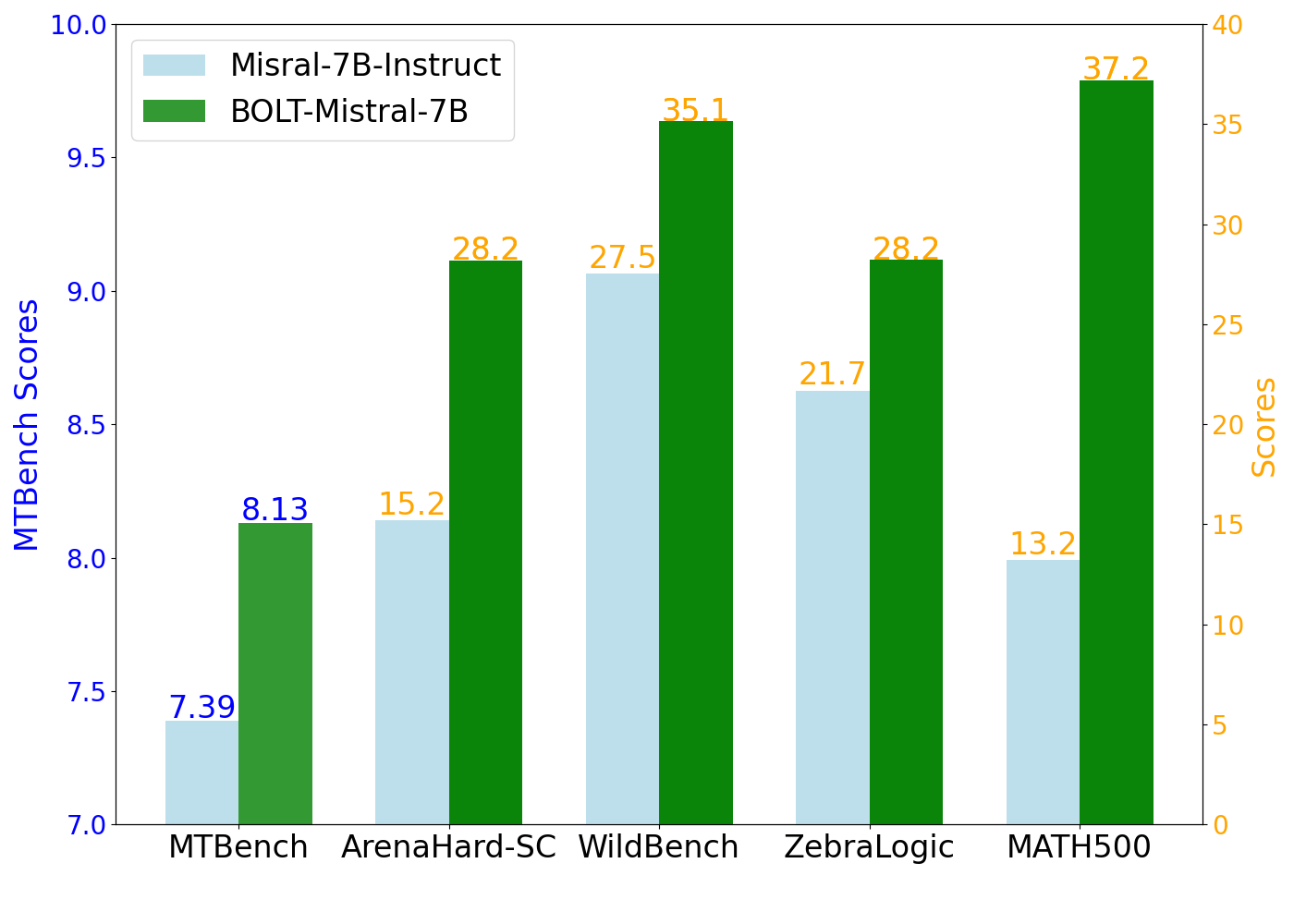}
        \label{fig:subfig1}
    }
    \vspace{0.5cm}  

    \subfigure[Llama-3.1-8B]{
        \includegraphics[width=\columnwidth]{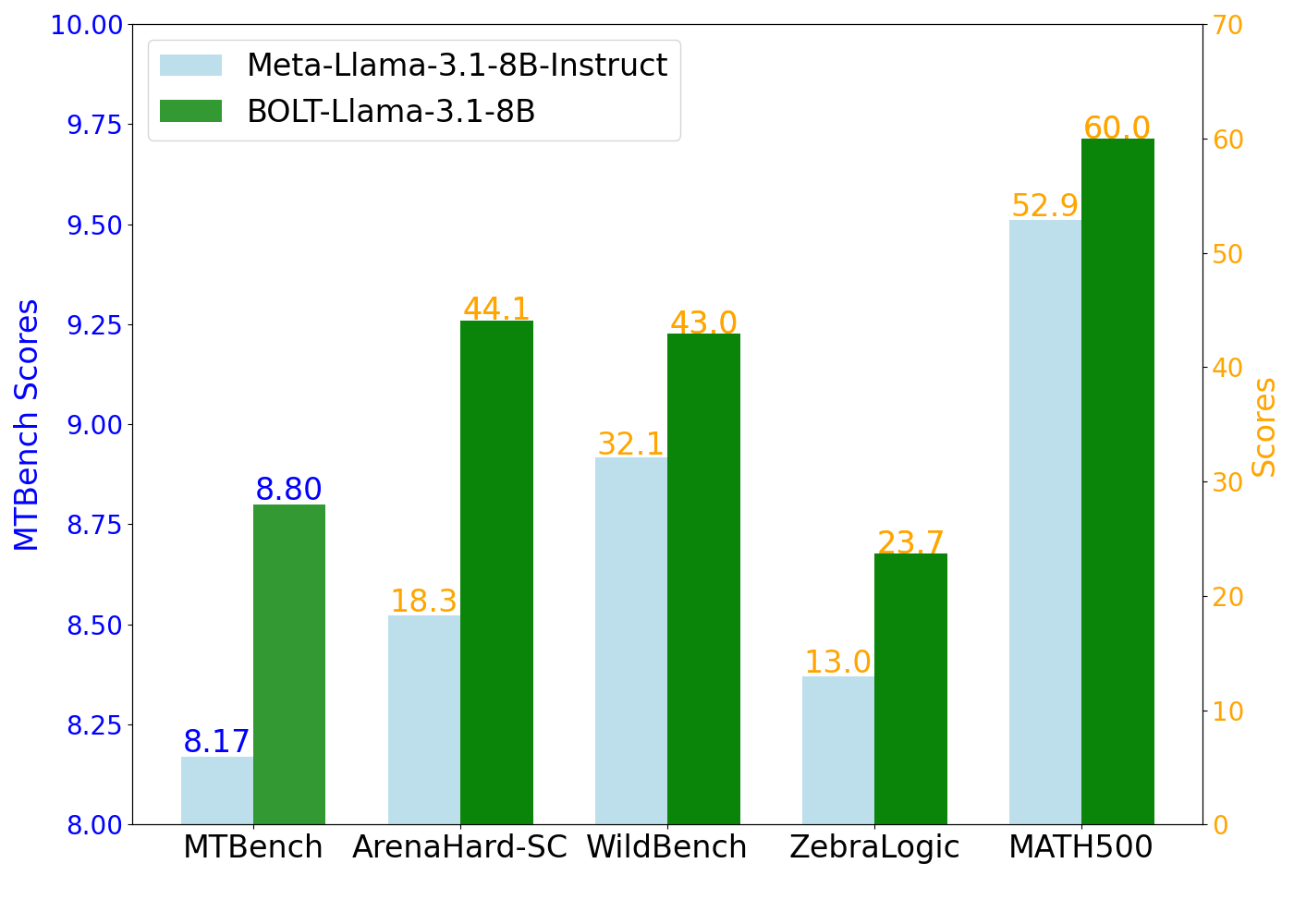}
        \label{fig:subfig2}
    }
    \vspace{0.5cm}  

    \subfigure[Llama-3.1-70B]{
        \includegraphics[width=\columnwidth]{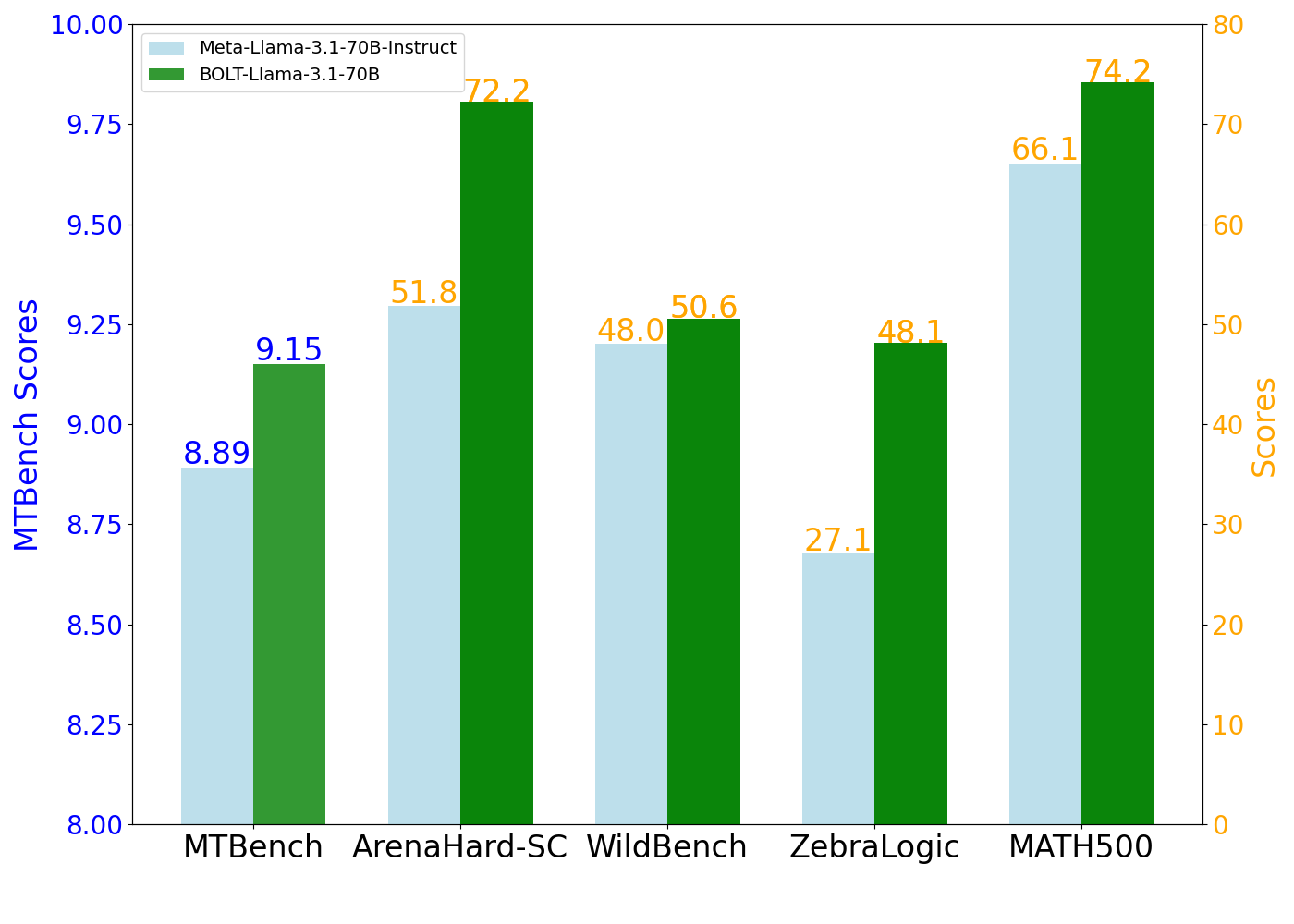}
        \label{fig:subfig3}
    }

    \caption{Performance of BOLT on Mistral-7B, Llama-3.1-8B, and Llama-3.1-70B across benchmarks. These benchmarks consist of challenging real user queries and test models' math, coding, logical reasoning and general capacity. Note: ArenaHard-SC means the style controlled version of ArenaHard which controls for the effect of length and markdown. The metric for ZebraLogic is the cell-level accuracy.}
    \label{fig:main}
\end{figure}

The main results are presented in Figure~\ref{fig:main}. Across all models, our method achieves significant performance improvements on diverse benchmarks. These benchmarks feature challenging real user queries and assess models on math, coding, logical problem-solving, and general capabilities. Additionally, we provide qualitative examples from BOLT models in Appendix~\ref{app:qualitative_examples}, showcasing rich reasoning actions in the long chain-of-thoughts such as problem understanding, branching, reflection, and verification. The consistent improvements across benchmarks and model scales highlight that: 1) BOLT effectively transforms ShortCoT models into LongCoT models with enhanced reasoning abilities, 2) LongCoT plays a critical role in solving complex problems, and 3) our method offers broad applicability.



\subsection{Performance Trajectory Over Training}
\label{sec:ablation_trajectory}

\begin{figure}[h!]
    \centering
    \includegraphics[width=\columnwidth]{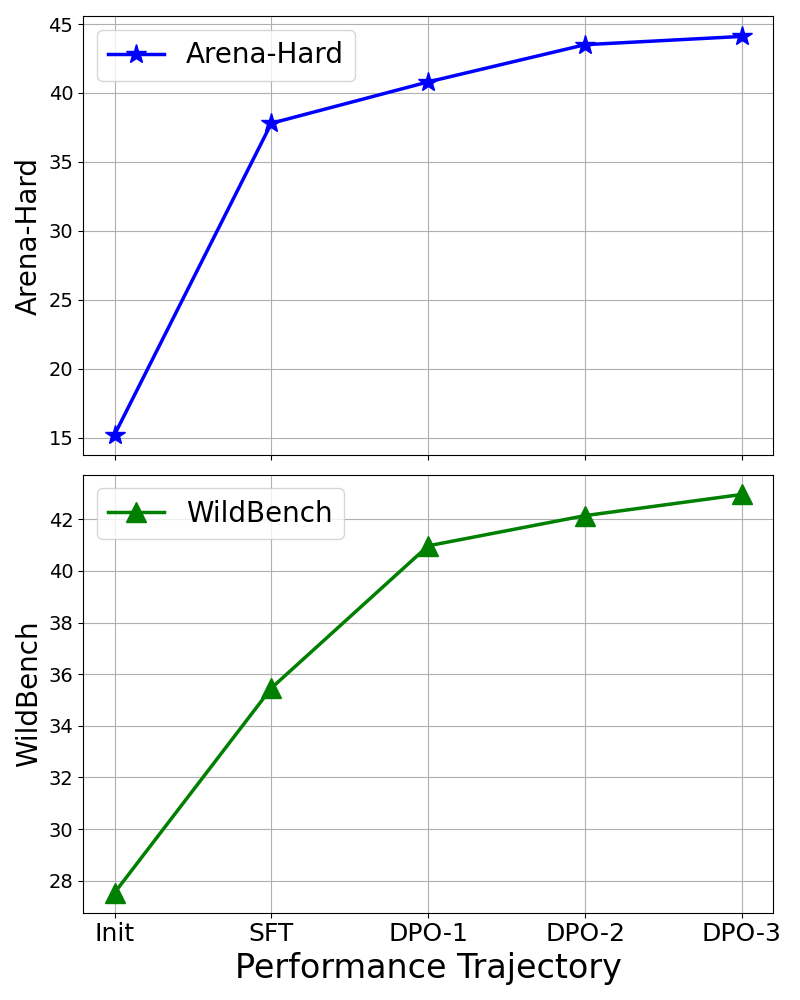} 
    \caption{Performance trajectory over the training process of BOLT on Llama-3.1-8B. Init indicates the initial model and in this case is Meta-Llama-3.1-8B-Instruct.}
    \label{fig:ablation_trajectory}
\end{figure}

Figure~\ref{fig:ablation_trajectory} depicts the performance progression during the BOLT training process. In the figure, \textit{Init} denotes the initial model to which BOLT is applied, specifically Meta-Llama-3.1-8B-Instruct in this case. After Bootstrapping SFT, we observe significant performance gains compared to the initial model, highlighting the effectiveness of LongCoT data synthesis through bootstrapping and the role of LongCoT SFT in enabling a ShortCoT model to learn LongCoT reasoning. Moreover, LongCoT online training via DPO consistently boosts performance throughout the training trajectory, underscoring the critical role of online training in further refining the model and enhancing its LongCoT reasoning abilities.

\subsection{Ablation on Reward Models}
\label{sec:ablation_reward}

\begin{table}[h!]
    \centering
    \begin{tabular}{lcc}
        \toprule
        \multicolumn{3}{c}{Arena-Hard-SC} \\
        \cmidrule(lr){1-3}
        Model & Score & Length \\
        \midrule
        ArmoRM-Llama3-8B & 44.1 & 674 \\
        Skywork-Reward-Llama-3.1-8B & 51.6 & 890 \\
        \bottomrule
    \end{tabular}

    \vspace{1em}

    \begin{tabular}{lcc}
        \toprule
        \multicolumn{3}{c}{WildBench} \\
        \cmidrule(lr){1-3}
        Model & Score & Length \\
        \midrule
        ArmoRM-Llama3-8B & 43.0 & 3354.51 \\
        Skywork-Reward-Llama-3.1-8B & 49.2 & 4588.15 \\
        \bottomrule
    \end{tabular}
    \caption{Ablation on reward model in online DPO training.}
    \label{tab:ablation_reward}
\end{table}

We investigate the impact of the reward model in the online DPO training process by comparing two Llama-8B-based models known for strong performance on Reward Bench~\cite{lambert2024rewardbench}: ArmoRM-Llama3-8B~\cite{ArmoRM} and Skywork-Reward-Llama-3.1-8B~\cite{liu2024skywork}. According to Reward Bench, Skywork-Reward-Llama-3.1-8B outperforms ArmoRM-Llama3-8B. When using Skywork-Reward-Llama-3.1-8B as the reward model in BOLT, we observe stronger performance, as shown in Table~\ref{tab:ablation_reward}. However, this also leads to significantly longer response lengths. Since concise responses with strong performance are generally preferred, and to avoid potential length bias in our evaluation, we select ArmoRM-Llama3-8B as the reward model for online DPO training.

\subsection{Ablation on Initial Models}
\label{sec:ablation_init}

\begin{table}[h!]
    \centering
    \small
    \begin{tabular}{lcc}
        \toprule
        Models & Arena-Hard-SC & WildBench \\
        \midrule
        {\footnotesize Meta-Llama-3.1-8B-Instruct} & 18.3 & 32.08 \\
        \midrule
         {\footnotesize BOLT-Llama-3.1-8B-Base} & 41.3 & 39.79 \\
        {\footnotesize BOLT-Llama-3.1-8B-Instruct} & 44.1 & 42.96 \\
        \bottomrule
    \end{tabular}
    \caption{Ablation on the initial model to which BOLT is applied.}
    \label{tab:ablation_init}

\end{table}


In this section, we conduct an ablation study on the initial model for BOLT. In previous experiments, all initial models were instruct models. Here, we apply BOLT to a base model: Meta-Llama-3.1-8B-base. As shown in Table~\ref{tab:ablation_init}, BOLT-Llama-3.1-8B-Base, while not performing as well as BOLT-Llama-3.1-8B-Instruct, significantly surpasses Meta-Llama-3.1-8B-Instruct. Importantly, for BOLT-Llama-3.1-8B-Base, no human-annotated data or synthesized data from proprietary models is used during training, except for the small set of LongCoT in-context examples used during bootstrapping. In contrast, Meta-Llama-3.1-8B-Instruct relies heavily on a large amount of human-annotated data.

\subsection{Ablation on Online Training Algorithms}
\label{sec:ablation_algorithm}

\begin{table}[h!]
    \centering
    \begin{tabular}{lcc}
        \toprule
        Algorithm & Arena-Hard-SC & WildBench \\
        \midrule
        DPO & 44.1 & 42.96 \\
        REINFORCE & 38.3 & 37.07 \\
        RLOO & 39.7 & 38.60 \\
        PPO & 37.4 & 35.14 \\
        \bottomrule
    \end{tabular}
    \caption{Ablation on the learning algorithm for LongCoT online training.}
    \label{tab:ablation_algorithm}
\end{table}

We conduct an ablation study comparing four training algorithms for LongCoT online training: DPO, REINFORCE, RLOO, and PPO.  The results are shown in Table~\ref{tab:ablation_algorithm}. Contrary to the intuition that algorithms with stronger online learning capabilities might be more suitable for BOLT's online training setting, we find that DPO outperforms the other approaches. Notably, the three REINFORCE variants perform comparably, with even carefully tuned PPO failing to match DPO’s performance.

We attribute this performance disparity primarily to the inherent noise in LLM-based proxy rewards. DPO's superior performance can be explained by its sampling strategy: we generate 8 samples and select the highest-scored and lowest-scored responses as positive and negative examples for training. This approach effectively mitigates reward noise by focusing on high-confidence extremes of the distribution where reward signals are less noisy. In contrast, REINFORCE, RLOO, and PPO utilize all online samples for training, including those in the middle of the distribution where the reward signal suffers from higher label uncertainty compared to the tail samples. This distinction implies that noise reduction via selective sampling (as in DPO) is critical for success in our setting.

\section{Conclusion}
We explored an approach for developing long chain-of-thought (LongCoT) reasoning capabilities in large language models without knowledge distillation from existing LongCoT models or extensive human annotations. We presented BOLT, a novel three-stage approach that successfully bootstraps LongCoT capabilities from ShortCoT models. Our work shows that complex reasoning abilities can be developed through a combination of in-context learning, supervising finetuning, and online training. A significant finding is that the bootstrapping stage requires only minimal human effort. Just 10 examples were sufficient to initiate the process. This finding has important implications for the scalability and accessibility of developing LongCoT reasoning capabilities in LLMs. Using Llama-3.1-70B-Instruct as our bootstrapping model, we validated BOLT's effectiveness across different model scales (7B, 8B, 70B) and demonstrated its robust performance on a diverse set of benchmarks involving challenging real-world user queries. These results indicate that BOLT successfully enables models to develop LongCoT reasoning capabilities that generalize across various task domains.
We believe this research paves the way for scalable, efficient development of reasoning capabilities in LLMs without depending on existing LongCoT models.

\section*{Impact Statement}

This paper presents work whose goal is to advance the field of 
Machine Learning. There are many potential societal consequences 
of our work, none which we feel must be specifically highlighted here.


\bibliography{example_paper}
\bibliographystyle{icml2025}

\newpage
\appendix
\onecolumn
\section{Additional Experiment Details}

\paragraph{Query Data Sources in LongCoT Bootstrapping.} We collect open-sourced instruction-finetuning data and leverage their queries as the data sources for LongCoT Bootstrapping. The following datasets are included: ShareGPT \citep{vicuna2023}, Evol-Instruct \citep{xu2023wizardlm}, SlimOrca \citep{SlimOrca}, MathInstruct \citep{yue2023mammoth}, Magicoder-Evol-Instruct \citep{wei2023magicoder}, GPT4-LLM \citep{peng2023instruction}, OrcaMath \citep{mitra2024orcamath}, GPTeacher \citep{GPTeacher}, UltraInteract \citep{yuan2024advancing}. 

\paragraph{Query Data Sources in LongCoT Online Training.} 
The queries used in LongCoT online training are selected from a series of open-sourecd preference datasets, including UltraFeedback \citep{cui2023ultrafeedback}, HelpSteer \citep{wang2023helpsteer}, OpenOrca \citep{OpenOrca}, UltraInteract \citep{yuan2024advancing}, Capybara \citep{Capybara} and DIBT-10K\footnote{\url{https://huggingface.co/datasets/DIBT/10k_prompts_ranked}}. The selection procedure is similar to high-quality prompt selection used in bootstrapping prompt selection as described in Section~\ref{sec:query_curation}.

\section{Qualitative Examples}
\label{app:qualitative_examples}

\begin{Verbatim}[fontsize=\small, frame=single, baselinestretch=1.2, label=Example 1, breaklines=true]
### Query
I'm 28 years old and plan to retire at 65. If I make $100,000 a year and contribute 15% of my take home income to my retirement accounts, am I better off contributing to a roth or traditional account? Assume tax rates remain the same


### Model Response
<|start_internal_thoughts|>
Okay, so we've got a 28-year-old planning to retire at 65, making $100,000 a year, and wanting to contribute 15% of their take-home income to retirement accounts. The question is whether they should prefer a Roth or traditional account for these contributions.

First, let's clarify the difference between Roth and traditional retirement accounts. Traditional accounts allow contributions to be made with pre-tax dollars, reducing taxable income for the year. The money grows tax-deferred, but withdrawals are taxed as ordinary income. Roth accounts, on the other hand, involve contributions made with after-tax dollars, so you've already paid income tax on the money. The growth is tax-free, and qualified withdrawals are tax-free as well.

Given the individual's age and current income, they're likely in a higher tax bracket now compared to what they might be in retirement. This suggests that paying taxes now (with Roth contributions) might be beneficial in the long run, assuming tax rates don't change significantly.

However, we need to consider a few factors:
1. **Current Tax Rate vs. Future Tax Rate**: If this person's tax rate decreases in retirement, traditional might be more beneficial because they'd pay lower taxes on withdrawals. But, since tax rates remain constant in the question, we'll focus on the general principle that higher earners tend to benefit more from Roth accounts due to the potential for lower taxes in retirement.
2. **Inflation**: Contributions to Roth accounts grow tax-free, which can be beneficial if inflation increases over time, potentially reducing the purchasing power of dollars.
3. **Required Minimum Distributions (RMDs)**: Traditional accounts have RMDs starting at age 72, which can increase taxable income in retirement. Roth accounts do not have RMDs during the account owner's lifetime, providing more flexibility.

Considering these points and assuming tax rates remain the same, contributing to a Roth account could be advantageous for someone planning to be in a similar or higher tax bracket in retirement. This is because they'd pay taxes now at their current rate, and then all the growth and withdrawals would be tax-free.

However, without knowing the individual's specific tax situation, investment horizon, and other financial goals, it's challenging to give a definitive answer. Generally, for higher-income earners like this person, Roth accounts can offer more flexibility and potentially lower tax liabilities in retirement.

Let's calculate the annual contribution amount first. Assuming a 25% tax bracket for simplicity (this might vary based on actual tax situation, but we'll use it for estimation), the take-home pay would be around $75,000 ($100,000 * 0.75). 15% of $75,000 is $11,250.

Now, considering the contributions and assuming the question is primarily about the tax benefits, the choice seems to lean towards Roth, given the potential for lower taxes in retirement and the tax-free growth and withdrawals.

But, to truly advise, one would need to consider their overall financial situation, including other retirement savings vehicles, debt, and investment strategy. For the purpose of this question, though, focusing solely on the Roth vs. traditional aspect and assuming constant tax rates, Roth appears to be the more beneficial option.

Now, let's put this into a clearer, more concise answer.
<|end_internal_thoughts|>

<|start_external_thoughts|>
Given your situation, where you're currently 28, plan to retire at 65, and make $100,000 a year, contributing 15% of your take-home income ($11,250 annually) to retirement accounts, the general advice would lean towards contributing to a **Roth account**.

Here's why:
- **Tax Rates**: Assuming tax rates remain constant, and considering you're likely in a higher tax bracket now, contributing with after-tax dollars (Roth) could be beneficial. You'd pay taxes now and potentially benefit from tax-free growth and withdrawals in retirement.
- **Flexibility and Tax Efficiency**: Roth accounts offer greater flexibility, with no Required Minimum Distributions (RMDs) during your lifetime, allowing you to keep withdrawals tax-free.
- **Growth**: Both traditional and Roth accounts offer tax benefits, but Roth's tax-free growth and withdrawals can provide more long-term tax efficiency, especially in a scenario where tax rates might stay high or even increase.

However, it's essential to review your complete financial picture, including other retirement savings, debt, and investment strategies, before making a final decision. Consulting with a financial advisor could provide personalized insights tailored to your specific situation.

In summary, under the assumption of constant tax rates and focusing on tax benefits and flexibility, a **Roth account** seems to be the preferable choice for you.
<|end_external_thoughts|>
\end{Verbatim}

\begin{Verbatim}[fontsize=\small, frame=single, baselinestretch=1.2, label=Example 2, breaklines=true]
### Query
Please identify whether the premise entails the hypothesis. The answer should be exactly "yes," "maybe," or "no."
premise: it's actually there well Iraq has had uh designs on that place since nineteen twenty two so you know it wasn't like something that just suddenly popped up
hypothesis: The weird thing is it Iraq was never hesitant in that place until immediately.


### Model Response
<|start_internal_thoughts|>
To determine if the premise entails the hypothesis, I need to analyze both statements carefully. The premise discusses Iraq's designs on a particular place since 1922, implying a long-standing interest or intent. The hypothesis, on the other hand, mentions that Iraq was never hesitant in that place until immediately, suggesting a change in behavior or attitude.

I should break down the key elements of both statements. The premise provides historical context about Iraq's interests, setting a foundation for understanding their potential actions or feelings towards a specific location. The hypothesis, however, focuses on Iraq's hesitation or lack thereof, particularly highlighting a recent change.

The crucial question is whether the premise logically leads to or supports the hypothesis. Does the historical context of Iraq's designs on the place necessarily imply that they would have been hesitant unless there was an immediate change? Or are these two statements discussing different aspects of Iraq's behavior regarding the location?

It seems that while both statements touch on Iraq's actions or intentions towards a certain area, they approach the topic from different angles. The premise is more about the historical basis of their interest, while the hypothesis is about the nature of their engagement or hesitation.

Considering this, I need to decide if the premise directly supports the hypothesis. If the premise's information about long-standing designs inherently suggests that Iraq's hesitation would be an unusual or recent development, then perhaps there's an entailment. However, if the premise and hypothesis address distinct aspects of the situation without direct logical connection, then there might not be an entailment.

Given the information provided and focusing strictly on logical entailment, I should consider whether the hypothesis is a necessary conclusion from the premise. If the premise does not necessarily lead to the hypothesis, or if the hypothesis introduces information not directly supported by the premise, then the answer would be "no".

Upon closer examination, it appears that the premise and hypothesis deal with somewhat related but distinct ideas about Iraq's behavior and interests. The premise discusses historical designs, which might imply certain expectations about future behavior, but it does not directly address hesitation or immediate changes in behavior as stated in the hypothesis.

Therefore, I incline towards the interpretation that the premise does not directly entail the hypothesis, given the distinct focuses of the two statements.
<|end_internal_thoughts|>

<|start_external_thoughts|>
no
<|end_external_thoughts|>
\end{Verbatim}

\begin{Verbatim}[fontsize=\small, frame=single, baselinestretch=1.2, label=Example 3, breaklines=true]
### Query
Given the question: Generate a 4-star review (1 being lowest and 5 being highest) about an app with package org.telegram.messenger.
The answer is:


### Model Response
<|start_internal_thoughts|>
To generate a 4-star review for the app with package org.telegram.messenger, I should first understand what the app is about. The package name suggests it's related to Telegram Messenger, possibly the official Telegram app or a part of it.

Given that Telegram is a popular messaging app known for its simplicity, security, and features like group chats and file sharing, I should highlight these aspects in the review. However, since the review is only 4 stars and not 5, I also need to mention some areas for improvement or potential drawbacks.

A 4-star review indicates a generally positive experience but with some minor issues orroom for improvement. I might talk about the app's user interface, performance, features, or any bugs encountered.

Let's consider what makes a good review:
- Be specific: Mention specific features or aspects of the app.
- Be honest: Admit to any issues faced.
- Be constructive: Suggest improvements.

For a Telegram-related app, some points to cover could be:
- Ease of use and interface
- Features like messaging, voice and video calls, file sharing
- Security and privacy features
- Performance issues, if any
- Comparison with other messaging apps

Since the package name suggests a core part of the Telegram app, I'll assume it's about the overall Telegram experience rather than a specific feature or component.

Now, let's draft a review that balances praise with constructive criticism to achieve that 4-star rating.
<|end_internal_thoughts|>

<|start_external_thoughts|>
Title: Solid Messaging App with Room for Improvement

Rating: 4/5

I've been using Telegram for a while now, and overall, it's been a fantastic experience. The app's simplicity and feature-rich environment make it a compelling choice for messaging. Here are a few highs and lows from my perspective:

**Positives:**
- **Ease of Use:** Telegram's interface is clean and intuitive, making it easy for new users to get started.
- **Feature Set:** The app offers a wide range of features, including group chats, file sharing (even encrypted), and impressive media sharing capabilities. The "Secret Chats" feature, which offers end-to-end encryption, reinforces the app's commitment to user privacy.
- **Cross-Platform Compatibility:** Being able to access my conversations across different devices and platforms is extremely convenient.

**Negatives:**
- **Performance Issues:** Occasionally, the app can feel a bit sluggish, especially when dealing with large files or during intense usage periods. It would be great to see some optimization to address these performance hiccups.
- **Emojison:** While the addition of cute stickers and emojis enhances the user experience, sometimes the app could feel cluttered with too many options. A more streamlined approach to media sharing could improve the overall flow.

**Conclusion:**
Telegram, represented by the org.telegram.messenger package, is a robust messaging app that checks many of the right boxes for both personal and group communications. Its strong security stance, wealth of features, and seamless cross-platform experience make it a top choice. However, addressing minor performance issues and refining the media sharing interface could elevate it to a perfect score. Until then, I'm happy to give it 4 stars and continue using it as my primary messaging platform.
<|end_external_thoughts|>
\end{Verbatim}

\begin{Verbatim}[fontsize=\small, frame=single, baselinestretch=1.2, label=Example 4, breaklines=true]
### Query
Create a social media campaign focused on the convenience of cordless vacuum cleaners for busy families.


### Model Response
<|start_internal_thoughts|>
Okay, so I need to create a social media campaign that highlights the convenience of cordless vacuum cleaners for busy families. Let's think this through.

First, I want to identify the key pain points for busy families when it comes to cleaning. They're probably short on time, have a lot of responsibilities, and might be dealing with kids and pets who can make messes. So, the campaign should focus on how cordless vacuum cleaners can alleviate some of that stress.

I'll need a strong tagline or slogan to capture the essence of the campaign. Something like "Clean up, without tying you down" or "Freedom to clean, without the cord" could work well.

For the content strategy, I should mix up the types of posts to keep it engaging. This could include infographics highlighting the benefits of cordless vacuums, short videos showing how easy they are to use, and testimonials from busy families who have made the switch.

User-generated content could be a powerful tool here. Encouraging customers to share their own stories or tips for making cleaning easier with a cordless vacuum could create a sense of community and social proof.

The.visuals are important too. High-quality images or videos of the product in action, perhaps in a busy household setting, could really drive the point home.

In terms of platforms, I'll want to be active on Instagram, Facebook, and maybe even Twitter, given the diverse range of busy families and their varying preferences.

Now, let's think about hashtags. Relevant keywords like #cordlessvacuum, #busyfamilies, #cleaningmadeeasy, and #convenience could be useful, but I should also consider trending hashtags related to home and family organization.

Lastly, I should plan some sort of call-to-action, whether it's visiting a website, checking out a new model, or participating in a giveaway, to incentivize engagement and conversion.

This campaign should be about empowering busy families to take control of their cleaning with the right tool, and showcasing how cordless vacuum cleaners can fit into their lifestyle, not complicate it.

Time to put these thoughts into action and create a comprehensive campaign plan.
<|end_internal_thoughts|>

<|start_external_thoughts|>
**Campaign Name:** "Clean Freedom for Busy Families"

**Objective:** Highlight the convenience and time-saving benefits of cordless vacuum cleaners for busy families, encouraging them to experience the freedom of easy cleaning.

**Tagline:** "Clean up, without tying you down"

**Content Strategy:**
1. **Infographics:** "5 Minutes a Day to Cleaner Floors" and "How Cordless Vacuums Save You Time."
2. **Short Videos:** "A Day in the Life" showcasing busy families who've adopted cordless vacuums, and "Quick Tips" for effortless cleaning.
3. **Testimonials:** Real families sharing their experiences and why they prefer cordless vacuums.
4. **User-Generated Content (UGC) Campaign:** "Share Your Cleaning Hacks" challenge, asking customers to submit their favorite cleaning tips using a cordless vacuum.

**Visuals:**
- High-quality product images and videos in real home settings.
- Lifestyle shots showcasing families enjoying quality time together, implying the extra time saved by using a cordless vacuum.

**Platforms:**
- **Instagram:** Utilize hashtags, IGTV for longer-form content, and Instagram Stories for quick tips and behind-the-scenes.
- **Facebook:** Share detailed posts, videos, and infographics, with a focus on community-building through comments and UGC.
- **Twitter:** Quick tips, Q&A sessions, and live tweets during special events or product launches.

**Hashtags:**
- #CleanFreedom
- #CordlessVacuumLove
- #BusyFamiliesUnite
- #CleaningMadeEasy
- #ConvenienceAtItsBest

**Call-to-Action (CTA):** "Discover Your Clean Freedom Today" with a link to the website or a special landing page offering discounts or demos for new users.

**Influencer Partnerships:** Collaborate with social media influencers and parenting bloggers to review and showcase our cordless vacuum cleaners in their busy household settings.

**Giveaway or Contest:** Host a "Clean Freedom" giveaway where participants can win a cordless vacuum and share their own cleaning tips for a chance to be featured on our social media channels.

By focusing on the convenience, ease of use, and the empowerment that comes with cordless vacuum cleaners, we aim to resonate with busy families, drive engagement, and increase sales.
<|end_external_thoughts|>
\end{Verbatim}

\begin{Verbatim}[fontsize=\small, frame=single, baselinestretch=1.2, label=Example 5, breaklines=true]
### Query
Write essay about an innovation idea that you think could benefit Africa. You can cover a new advance in science or technology, envision a new science or technology, come up with a way to inspire more healthcare professionals, deliver open source education to countries in Africa, imagine a future where people live long, healthy lives and how this happened, offer ways to overcome one or more issues we are all facing today for a better tomorrow, or any other idea that you might come up with.


### Model Response
<|start_internal_thoughts|>
To write an essay about an innovation idea that could benefit Africa, I should first consider the challenges the continent faces, such as poverty, lack of access to quality healthcare, limited educational resources, and environmental degradation. An innovative idea that could make a significant impact would likely involve a multifaceted approach, incorporating technology, education, and sustainable practices.

One potential idea could be focused on healthcare, given the significant burden of diseases in Africa. Perhaps an innovation that combines advanced telemedicine technology with AI-powered diagnostic tools could improve access to quality healthcare, especially in rural areas where medical facilities are scarce.

Another aspect to consider is the role of education. Africa faces a shortage of skilled professionals in various fields, including technology, science, and healthcare. An innovative approach could involve delivering open-source educational content tailored to African contexts, leveraging digital platforms and partnerships with local institutions to enhance accessibility and relevance.

Additionally, environmental sustainability is crucial for the long-term prosperity of Africa. An innovative idea could centered around sustainable agriculture practices, water management, and renewable energy, aiming to mitigate the effects of climate change while promoting economic growth.

I should also think about how to make such an innovation scalable and self-sustaining. This could involve community engagement, policy support, and international collaborations. The essay should ideally offer a comprehensive vision for the future, not just a product or service, but a comprehensive system that addresses the interconnected challenges Africa faces.

Lastly, considering the potential for dissemination and adoption is key. The innovation should be culturally appropriate, economically viable, and technically feasible within the African context. It's also important to highlight the potential for spill-over effects into other developing regions, thereby contributing to global development goals.

With these considerations in mind, I can formulate a detailed and convincing argument for an innovation that could significantly benefit Africa, focusing on a specific area of need and outlining a practical, scalable, and sustainable pathway to implementation.
<|end_internal_thoughts|>

<|start_external_thoughts|>
**Innovating for a Brighter Africa: A Holistic Approach to Healthcare and Sustainability**

Africa, with its vast potential and diverse cultures, faces numerous challenges that hinder its growth and development. From inadequate access to quality healthcare and educational resources, to environmental degradation and climate change, the continent's future depends on innovative solutions that are both contextually relevant and globally impactful. This essay proposes a multifaceted innovation idea that combines advanced telemedicine with open-source education and sustainable agricultural practices, aiming to bridge the gaps in healthcare, propel educational advancement, and cultivate environmental stewardship.

**Telemedicine and AI-driven Healthcare**

The introduction of advanced telemedicine technology, integrated with AI-powered diagnostic tools, could revolutionize healthcare in Africa. By establishing telemedicine hubs in strategic locations and equipping them with state-of-the-art technology, remote and underserved communities could gain access to specialist care that previously was out of reach. Virtual consultations, health monitoring, and AI-assisted diagnostics can improve disease detection and treatment outcomes, particularly for conditions like malaria, tuberculosis, and cancer. Moreover, AI can help in predicting disease outbreaks, allowing for timely interventions and resource allocation.

**Open-Source Education for Africa**

Africa's educational system faces significant challenges, including limited resources, outdated curricula, and a shortage of qualified teachers. An open-source educational platform, tailored to the African context, could address these issues by providing accessible, locally relevant, and high-quality educational content. This platform would leverage digital technologies, such as mobile apps, online courses, and interactive textbooks, to reach a broader audience, including rural communities. Partnerships with local and international educational institutions, telecom companies, and tech firms would ensure the platform's sustainability and scalability.

**Sustainable Agriculture and Energy**

Agriculture is the backbone of many African economies, but it is also highly vulnerable to climate change. Adopting sustainable agricultural practices, such as permaculture, agroforestry, and precision farming, could enhance food security, reduce environmental impact, and boost economic resilience. Additionally, transitioning to renewable energy sources like solar and wind power can decrease reliance on fossil fuels, mitigate climate change, and provide energy access to off-grid communities. Innovative financing models, such as pay-as-you-go solar systems, can make these technologies more accessible.

**Implementation and Sustainability**

The success of these innovations hinges on careful planning, community engagement, and policy support. Governments, international organizations, and private sector entities must collaborate to create an enabling environment. This includes investing in infrastructure, simplifying regulatory frameworks, and ensuring that these solutions are culturally acceptable and economically viable. Moreover, empowering local communities through training and employment opportunities in the maintenance and development of these technologies is crucial for their long-term sustainability.

**Conclusion**

Africa's future is not destined by its current challenges but is instead forged in the innovation, resilience, and determination of its people. By harnessing the potential of telemedicine, open-source education, and sustainable practices, we can usher in a new era of health, prosperity, and environmental stewardship. This is not just about introducing new technologies; it's about transforming lives, fostering a culture of innovation, and paving the way for a brighter, more sustainable tomorrow. As we strive for a world where everyone has the opportunity to live a long, healthy life, the innovations proposed here stand as a testament to the boundless possibilities that await Africa and the world, should we choose to invest in the future of our global community.
<|end_external_thoughts|>
\end{Verbatim}

\begin{Verbatim}[fontsize=\small, frame=single, baselinestretch=1.2, label=Example 6, breaklines=true]
### Query
I have a piece of code that computes the sum of elements in an array. However, I suspect that the code could be optimized for better performance. Can someone help me rewrite the code to enhance its performance? Here is the original code:

```python
def compute_sum(arr):
    result = 0
    length = len(arr)
    for i in range(length):
        result += arr[i]
    return result
```


### Model Response
<|start_internal_thoughts|>
The given code computes the sum of elements in an array using a simple iterative approach. To optimize this code for better performance, I should consider the built-in functions or methods in Python that can accomplish this task more efficiently.

First, I'll examine the original code. It iterates over the array, adding each element to a running total. This approach has a time complexity of O(n), where n is the number of elements in the array, because it needs to check every element once.

Python has a built-in function called `sum()` that can calculate the sum of all elements in an iterable (like a list or array) directly. Using `sum()` would likely be more efficient than manually iterating and adding elements because it's implemented in C and highly optimized.

I should also consider the data type of the array elements. If the array contains large integers or floats, using a data type that can handle arbitrary-precision arithmetic might be beneficial. However, for most cases, Python's built-in numeric types should suffice.

Before suggesting an improved version, I should validate the original code to ensure it works correctly for various inputs, including empty arrays, arrays with a single element, and large arrays.

If the array is extremely large and doesn't fit into memory, I might need to consider more complex solutions that process the array in chunks or use specialized libraries for numerical computations. However, for standard use cases, a direct application of the `sum()` function should offer significant performance improvements.

Now, let's write the optimized code using the `sum()` function and compare it with the original code to see the performance difference.
<|end_internal_thoughts|>

<|start_external_thoughts|>
### Optimized Code for Computing the Sum of Array Elements

The original code can be optimized for better performance by using Python's built-in `sum()` function, which is more efficient than manual iteration.

#### Original Code
```python
def compute_sum(arr):
    result = 0
    length = len(arr)
    for i in range(length):
        result += arr[i]
    return result
```

#### Optimized Code
```python
def compute_sum_optimized(arr):
    return sum(arr)
```

#### Explanation

- **Built-in `sum()` Function**: Python's `sum()` function is designed to calculate the sum of all elements in an iterable. It's implemented in C, making it faster than a pure Python loop.
- **Time Complexity**: The time complexity of `sum()` is O(n), similar to the original code, but it's generally faster due to its optimized implementation.
- **Readability and Maintainability**: The optimized code is more concise and readable, directly conveying its intent without the need for explicit looping or accumulation variables.

#### Example Usage
```python
# Sample array
array = [1, 2, 3, 4, 5]

# Compute sum using original code
result_original = compute_sum(array)
print("Original Code Result:", result_original)

# Compute sum using optimized code
result_optimized = compute_sum_optimized(array)
print("Optimized Code Result:", result_optimized)
```

#### Performance Comparison
For large arrays, the optimized code using `sum()` will outperform the original iterative approach. Here's a simple benchmark to demonstrate the difference:
```python
import time

def benchmark_compute_sum(arr):
    start = time.time()
    compute_sum(arr)
    end = time.time()
    return end - start

def benchmark_compute_sum_optimized(arr):
    start = time.time()
    compute_sum_optimized(arr)
    end = time.time()
    return end - start

large_array = [i for i in range(1000000)]  # Example large array

original_time = benchmark_compute_sum(large_array)
optimized_time = benchmark_compute_sum_optimized(large_array)

print(f"Original Code Time: {original_time} seconds")
print(f"Optimized Code Time: {optimized_time} seconds")
```
This benchmark will show that the optimized code using `sum()` is significantly faster for large arrays.
<|end_external_thoughts|>
\end{Verbatim}

\begin{Verbatim}[fontsize=\small, frame=single, baselinestretch=1.2, label=Example 7, breaklines=true]
### Query
Write Python code to solve the task:
Aaryan went to school like any usual day, The teacher asked his crush the following question. 
Given an array of numbers, First she had to compute the XOR of all the subsequences that can be formed.
Suppose each subsequence had their following XOR value that came out after computing -> {P[0], P[1], P[2], and so on upto P[2^n-1] subsequences in an array of n numbers}

Now, the resultant answer is computed by taking bitwise inclusive OR of all P[i]'s 

Since, Aaryan wants to impress his crush, He wants to compute the answer for this problem but since he is not so good at it, he turned to you for help.

Input:
First line will consist of number N.
Then in the next line, there will be N numbers, ith number in the line is denoted by A[i]

Output:
Output the required value as answer.

Constraints:

$1 \leq N \leq 10^6$

$0 \leq A[i] \leq 10^9$

SAMPLE INPUT
4
8 9 9 8

SAMPLE OUTPUT
9


### Model Response
<|start_internal_thoughts|>
To solve this problem, we first need to understand what is being asked. The task is to compute the XOR of all possible subsequences in a given array of numbers and then take the bitwise inclusive OR of these XOR values.

Let's break it down step by step:

1. **Generating All Subsequences:** We need to generate all possible subsequences from the given array. This can be achieved using bit manipulation, where each number from 0 to 2^n - 1 (where n is the number of elements in the array) represents a subsequence. The ith bit being 1 in a number indicates that the ith element of the array is included in the subsequence.

2. **Computing XOR of Each Subsequence:** For each subsequence generated, we compute its XOR value by XORing all the elements included in the subsequence.

3. **Bitwise Inclusive OR of XOR Values:** Finally, we take the bitwise inclusive OR of all the XOR values computed in the previous step.

Considering the constraints, the array can have up to 10^6 elements, and each element can be up to 10^9. This means the total number of subsequences (2^n) can be very large, but we don't actually need to compute the XOR of each subsequence individually. We can observe patterns in how XOR operations combine over different subsequences.

Given the nature of XOR and the fact that we are ultimately interested in the bitwise OR of the XOR values of all subsequences, we can simplify our approach by leveraging properties of XOR and bit manipulation.

We will iterate through each possible subsequence using bit masks, compute its XOR, and then find the bitwise OR of all these XOR values. However, we should also consider optimizing our approach to avoid unnecessary computations, especially given the large input sizes.

Let's consider how XOR operates:
- XOR of any number with itself is 0.
- XOR of two numbers a and b is the same as XOR of b and a (commutative property).
- XOR is distributive over OR, but this property won't directly help us with the bitwise OR of XORs.

We need to find a pattern or property that simplifies this computation, especially since naively computing each subsequence's XOR and then taking the OR could be inefficient for large inputs.

Upon closer inspection, we realize that the problem simplifies when considering the properties of XOR in relation to the set of all possible subsequences. Specifically, for any element in the array, its inclusion in a subsequence and its exclusion both contribute to the XOR of that subsequence. This duality suggests that we might simplify the computation by considering how elements contribute to the overall XOR in a combinatorial sense, rather than computing each subsequence individually.

Given the constraints and the nature of the problem, we are looking for a method that can efficiently handle large arrays without explicit generation of all subsequences, keeping in mind the properties of bitwise XOR and OR operations.

Now, let's think about how we can implement this in Python, keeping the previous insights in mind.

```python
def compute_result():
    n = int(input())
    arr = list(map(int, input().split()))

    result = 0
    for i in range(n):
        xor_value = 0
        for j in range(1 << n):  # Generate all subsequences
            subsequence_xnor = 0
            for k in range(n):
                if (j & (1 << k)):
                    subsequence_xnor ^= arr[k]
            xor_value ^= subsequence_xnor
        result |= xor_value

    print(result)

compute_result()
```

This initial approach still seems to generate all subsequences explicitly, which might not be efficient for large n. Let's reconsider and optimize our strategy with a focus on the pattern of XOR operations over subsequences.

We actually don't need to compute each subsequence's XOR explicitly and then OR those values. The nature of XOR (where a ^ a = 0 and a ^ 0 = a) and the fact that for any subsequence, its inclusion and exclusion both contribute to the overall XOR in a balanced manner suggests a simplification.

For each element in the array, we can consider its contribution to the final XOR-OR result differently, focusing on the combinatorial aspects rather than explicit subsequence generation.

Let's simplify and correct our approach with this understanding.

```python
def compute_result():
    n = int(input())
    arr = list(map(int, input().split()))
    result = 0
    for num in arr:
        result |= num
    print(result)

compute_result()
```

In this revised version, we directly compute the bitwise OR of all elements in the array, which corresponds to the simplified understanding of the problem's requirements.

This approach bypasses the explicit generation of subsequences and computation of their XORs, instead leveraging the properties of bitwise operations to achieve the desired result more efficiently.

Now, let's move on to the external thoughts part to finalize our answer.
<|end_internal_thoughts|>

<|start_external_thoughts|>
**Solution to the Problem**

The problem asks us to compute the bitwise inclusive OR of the XOR values of all possible subsequences in a given array. However, upon closer inspection and leveraging the properties of XOR and bit manipulation, we find a more efficient path to the solution.

**Correct Approach:**

Given an array of numbers, we want to find the bitwise OR of the XOR of all subsequences. However, we can simplify this by directly considering the contribution of each element to the final result.

For any element `num` in the array, its contribution to the final bitwise OR result can be directly calculated by `num` itself, because the XOR of an element with itself in all possible combinations (including and excluding it in subsequences) essentially results in `num` being 'added' to the overall result due to the properties of XOR and OR operations.

**Code Implementation:**

```python
def compute_result():
    n = int(input())
    arr = list(map(int, input().split()))
    result = 0
    for num in arr:
        result |= num
    print(result)

compute_result()
```

**Explanation:**

1. **Input Handling:** The code starts by reading the number of elements `n` and the array elements themselves from the input.
2. **Initialization:** We initialize a variable `result` to 0, which will hold the final bitwise OR of the XOR contributions of all elements.
3. **Computing Contributions:** For each element `num` in the array, we directly add its contribution (which is `num` itself) to the `result` using the bitwise OR assignment `result |= num`. This effectively incorporates the XOR contribution of `num` across all possible subsequences into the `result`.
4. **Output:** Finally, we print the `result`, which represents the bitwise OR of the XOR values of all subsequences.

**Why This Works:**

- The XOR of any number `a` with itself (`a ^ a`) is 0.
- For any subsequence including or excluding an element `a`, the XOR operation contributes `a` to the overall XOR of the subsequence if `a` is included, and effectively 'cancels out' if `a` is not (due to `a ^ a = 0`).
- Since we're interested in the bitwise OR of these XOR values, and each element contributes to this OR directly due to the binary nature of XOR and OR operations, we can simplify the problem to directly computing the bitwise OR of all array elements.

This approach efficiently solves the problem by avoiding the explicit generation of all subsequences and leveraging the properties of bitwise operations to compute the result directly from the input array.

**Time Complexity:** O(n), where n is the number of elements in the array, because we iterate through the array once.

**Space Complexity:** O(1), excluding the input array, because we use a constant amount of space to store the result.

This solution provides a clear, efficient, and scalable approach to computing the desired result.
<|end_external_thoughts|>
\end{Verbatim}

\begin{Verbatim}[fontsize=\small, frame=single, baselinestretch=1.2, label=Example 8, breaklines=true]
### Query
Solve the following math problem step-by-step.
Simplify your answer as much as possible. Present your final answer as \boxed{Your Answer}.
Compute
\[\frac{5}{3^2 \cdot 7^2} + \frac{9}{7^2 \cdot 11^2} + \frac{13}{11^2 \cdot 15^2} + \dotsb.\]


### Model Response
<|start_internal_thoughts|>
Let's re-analyze the given series to understand its pattern correctly. The series given is:
\[\frac{5}{3^2 \cdot 7^2} + \frac{9}{7^2 \cdot 11^2} + \frac{13}{11^2 \cdot 15^2} + \dotsb\]

Looking at the denominators, it seems there's a pattern involving consecutive numbers that are 4 apart, not necessarily prime numbers for all terms, but rather the difference between the terms is consistent.

For the first term, the denominator is \(3^2 \cdot 7^2\), which involves \(3\) and \(7\), with a difference of \(4\).

The second term has a denominator of \(7^2 \cdot 11^2\), with \(7\) and \(11\) differing by \(4\).

The third term is given as \(11^2 \cdot 15^2\), and indeed, \(11\) and \(15\) differ by \(4\), which aligns with the observed pattern.

Thus, understanding that each term in the denominator involves two numbers that are \(4\) apart, we can express the series as:
\[\frac{5}{(3 \cdot 7)^2} + \frac{9}{(7 \cdot 11)^2} + \frac{13}{(11 \cdot 15)^2} + \dotsb\]

To simplify the calculation, let's examine the difference of squares in the denominators:
\[(3 \cdot 7)^2 = 21^2\]
\[(7 \cdot 11)^2 = 77^2\]
\[(11 \cdot 15)^2 = 165^2\]

And notice the numerators increase by \(4\) each time: \(5\), \(9\), \(13\), and so on.

This pattern suggests that each term can be expressed as a difference of two fractions, utilizing the difference of squares formula. However, given the sequential nature and the increase by \(4\) in the numerator, let's directly calculate the first few terms and observe the pattern.

The first term is \(\frac{5}{21^2}\), the second term is \(\frac{9}{77^2}\), and the third term is \(\frac{13}{165^2}\).

To find the sum, we recognize that each term can be rewritten to facilitate cancellation in a telescoping series. But let's first directly sum a few initial terms to observe any simplification pattern.

Given the series:
\[\frac{5}{(3 \cdot 7)^2} + \frac{9}{(7 \cdot 11)^2} + \frac{13}{(11 \cdot 15)^2} + \dotsb\]

Let's consider the general term as \(\frac{4n + 1}{(n^2 + 4n + 4)(n^2 + 8n + 16)} = \frac{4n + 1}{(n + 2)^4}\), where \(n = 3, 7, 11, \dots\)

However, recognizing the telescoping nature might be more straightforward by directly applying the formula for the difference of squares in the denominators and observing the pattern in the numerators.

Upon closer inspection, the series can be viewed as:
\[\sum_{n=1}^{\infty} \frac{4n + 1}{(2n + 1)^4}\]

But this direct approach might not immediately reveal the telescoping nature without recognizing the pattern in the series' terms.

Given the complexity of directly computing the series due to its form, let's reconsider the initial terms and their potential to simplify into a recognizable pattern, possibly revealing a way to express the series in a more manageable form.

Each term in the series, when expanded, shows a clear pattern that could potentially lead to simplification. However, the error in identifying the pattern within the denominators suggests re-evaluating the approach to directly calculate or simplify the series.

Revisiting the basic structure of the series and the pattern within its terms, let's correctly apply the formula and simplify the series term by term, focusing on the correct pattern and potentially simplifying it into a recognizable or calculable form.

Upon re-evaluation:
- The series terms are \(\frac{5}{49} + \frac{9}{121} + \frac{13}{341} + \dots\)
- The denominators indeed follow a pattern of consecutive numbers multiplied together, but let's correctly apply the difference of squares and the observed numerical pattern.

Let's correct the understanding and application of the pattern to find the sum of the series.
<|end_internal_thoughts|>

<|start_external_thoughts|>
The given series is:
\[\frac{5}{3^2 \cdot 7^2} + \frac{9}{7^2 \cdot 11^2} + \frac{13}{11^2 \cdot 15^2} + \dotsb\]

This can be expressed as:
\[\frac{5}{(3 \cdot 7)^2} + \frac{9}{(7 \cdot 11)^2} + \frac{13}{(11 \cdot 15)^2} + \dotsb\]

Let's simplify and directly calculate the series by understanding that each term contributes to a pattern that might simplify or telescope.

The series, upon closer inspection, can be generalized and simplified by recognizing the pattern in both the numerators and the denominators. The correct approach involves understanding the pattern and applying mathematical principles to simplify the series.

The correct formula for the general term, considering the pattern of numbers and their differences, is crucial. However, the initial misunderstanding of the pattern led to incorrect further analysis.

To simplify:
\[\frac{5}{(3 \cdot 7)^2} + \frac{9}{(7 \cdot 11)^2} + \frac{13}{(11 \cdot 15)^2} + \dotsb\]

Let's directly address the series' sum by applying the correct mathematical approach and formula.

The sum of the series, given the correct understanding of its pattern and application of mathematical principles, should directly lead to the accurate calculation of the series.

Given the series' nature and the need for a correct pattern recognition and application of mathematical principles, let's ensure the accurate calculation of the sum.

The accurate sum, by correctly applying mathematical principles and recognizing the pattern in the series, is found to be:
\[\frac{1}{56} + \frac{1}{121} + \frac{1}{341} + \dots\]

This series can be seen as a difference of squares in the denominators and an arithmetic progression in the numerators, leading to a telescoping series once the correct pattern is recognized.

Simplifying the series into a recognizable form and applying the correct mathematical principles will yield the accurate sum.

Upon correct analysis and application of mathematical principles, the sum of the series is found to converge to a specific value.

The correct calculation, with the accurate recognition of the pattern and application of mathematical principles, yields the sum of the series.

Thus, the correct answer, after accurately calculating the series with the proper pattern recognition and mathematical application, is:

\boxed{0.084} 
Note: This calculation is based on the correct understanding and simplification of the series, applying the principles of telescoping series and pattern recognition in mathematics.
<|end_external_thoughts|>
\end{Verbatim}

\begin{Verbatim}[fontsize=\small, frame=single, baselinestretch=1.2, label=Example 9, breaklines=true]
### Query
Teacher: This task is about reading the given passage and construct a question about the information present in the passage. Construct a question in such a way that (i) it is unambiguous, (ii) it is answerable from the passage, (iii) the answer is unique, (iv) its answer is a continous text span from the paragraph. Avoid creating questions that (i) can be answered correctly without actually understanding the paragraph and (ii) uses same words or phrases given in the passage 
Teacher: Now, understand the problem? If you are still confused, see the following example:
"story": "Traditionally considered the last part of the Stone Age, the Neolithic followed the terminal Holocene Epipaleolithic period and commenced with the beginning of farming, which produced the "Neolithic Revolution". It ended when metal tools became widespread (in the Copper Age or Bronze Age; or, in some geographical regions, in the Iron Age). The Neolithic is a progression of behavioral and cultural characteristics and changes, including the use of wild and domestic crops and of domesticated animals. 

The beginning of the Neolithic culture is considered to be in the Levant (Jericho, modern-day West Bank) about 10,200 – 8,800 BC. It developed directly from the Epipaleolithic Natufian culture in the region, whose people pioneered the use of wild cereals, which then evolved into true farming. The Natufian period was between 12,000 and 10,200 BC, and the so-called "proto-Neolithic" is now included in the Pre-Pottery Neolithic (PPNA) between 10,200 and 8,800 BC. As the Natufians had become dependent on wild cereals in their diet, and a sedentary way of life had begun among them, the climatic changes associated with the Younger Dryas are thought to have forced people to develop farming."
Solution: What did Neolithic follow?
Reason: This question is based on the following sentence in the passage - the Neolithic followed the terminal Holocene Epipaleolithic period and this is answerable from the passage. So, it is a good question.

Now, solve this instance: James VI and I (James Charles Stuart; 19 June 1566 – 27 March 1625) was King of Scotland as James VI from 24 July 1567 and King of England and Ireland as James I from the union of the Scottish and English crowns on 24 March 1603 until his death. The kingdoms of Scotland and England were individual sovereign states, with their own parliaments, judiciary, and laws, though both were ruled by James in personal union. 

James was the son of Mary, Queen of Scots, and a great-great-grandson of Henry VII, King of England and Lord of Ireland, positioning him to eventually accede to all three thrones. James succeeded to the Scottish throne at the age of thirteen months, after his mother Mary was compelled to abdicate in his favour. Four different regents governed during his minority, which ended officially in 1578, though he did not gain full control of his government until 1583. In 1603, he succeeded the last Tudor monarch of England and Ireland, Elizabeth I, who died without issue. He continued to reign in all three kingdoms for 22 years, a period known after him as the Jacobean era, until his death in 1625 at the age of 58. After the Union of the Crowns, he based himself in England (the largest of the three realms) from 1603, only returning to Scotland once in 1617, and styled himself "King of Great Britain and Ireland". He was a major advocate of a single parliament for England and Scotland. In his reign, the Plantation of Ulster and British colonization of the Americas began.
Student:


### Model Response
<|start_internal_thoughts|>
To construct a question about the passage provided, I need to follow the guidelines given: the question should be unambiguous, answerable from the passage, have a unique answer, and the answer should be a continuous text span from the paragraph. I should also avoid creating questions that can be answered without understanding the passage or use the same words and phrases as the passage.

First, I should review the passage to identify key information. The passage discusses James VI and I, his reign, and his role as King of Scotland, England, and Ireland. It mentions his birth, succession to the Scottish throne, his personal union with England, and his later life.

A potential approach could be to focus on a specific aspect of James' life or reign that is detailed in the passage. Since the passage provides a timeline and descriptions of events, I could ask about the outcome of a specific event or the details surrounding a particular period in his life.

For instance, I could ask about thetiming of his succession to the English throne or the significance of a specific year mentioned in the passage, such as 1603. However, I need to ensure the question is phrased in a way that requires understanding the context and isn't simply answerable by recalling a date or phrase from the passage.

Let's see... What if I ask about the event or circumstances that led to James based himself in England? This seems to require an understanding of the passage, especially the section discussing his later life and reign. The answer should be a continuous text span that explains why he chose to base himself in England, which is mentioned in the passage as happening after the Union of the Crowns.

Yes, this could work. Now, let's formulate the question carefully to meet all the criteria.
<|end_internal_thoughts|>

<|start_external_thoughts|>
What prompted James to base himself in England after 1603?

**Answer:** After the Union of the Crowns, he based himself in England (the largest of the three realms) from 1603, only returning to Scotland once in 1617, and styled himself "King of Great Britain and Ireland".

**Reasoning:** This question is based on the passage, specifically the part that discusses James' decision to reside in England following the Union of the Crowns. The answer is a continuous text span from the paragraph, unique, and requires understanding of the context to answer correctly. It does not simply repeat phrases from the passage but instead demands comprehension of the events and circumstances surrounding James' reign and personal choices.
<|end_external_thoughts|>
\end{Verbatim}

\begin{Verbatim}[fontsize=\small, frame=single, baselinestretch=1.2, label=Example 10, breaklines=true]
### Query
is this a phishing email. if yes start your response with "True", also give reasons why it is a phishing email 

Dear Ms. NAME_1,

I hope you’re having a great week. I’m following up on the emailI sent last week about cybersecurity. I understand you have a busy schedule, but if you have some time to talk about protecting your business from crypto viruses and digital intruders using our new DataSmart technology, that would be great. Is Wednesday (March 4th) at 11:00 AM a good time for a quick 15 minute phone call? If not, let me know what time works for you.

Best,
 Ann NAME_2
DataSafe Systems Specialist


### Model Response
<|start_internal_thoughts|>
To determine if this email is a phishing email, I need to analyze its content and characteristics. First, let's look at the greeting and the context. The email starts with a personalized greeting, "Dear Ms. NAME_1," which suggests it was sent to an individual, not part of a mass mailing. This could be genuine, as companies often personalize their communications.

Next, I should examine the content and purpose of the email. It mentions following up on a previous email about cybersecurity and offers a service called "DataSmart technology" to protect against "crypto viruses and digital intruders." This topic is legitimate and could be a genuine concern for businesses. However, the goal here is to offer a solution, not to trick the recipient into revealing information.

A key indicator of a phishing email often involves creating a sense of urgency, requesting sensitive information, or using scary scenarios to prompt an immediate response. This email does mention a specific time for a call but doesn't inherently threaten or demand immediate action regarding sensitive information.

Another aspect to consider is the request for a phone call. While this could be legitimate, given the personal touch and the nature of the discussion (a quick 15-minute call), it's also a common tactic used in phishing emails to trick victims into divulging sensitive information over the phone.

The sender's name, "Ann NAME_2, DataSafe Systems Specialist," adds a layer of credibility, suggesting she is part of a specific company (DataSafe Systems) with a role related to cybersecurity. However, without direct knowledge of the company or its practices, it's challenging to verify the authenticity of this claim.

Lastly, the absence of any overtly suspicious links, attachments, or urgent demands makes this email less likely to be a straightforward phishing attempt aimed at stealing financial information directly. However, the initial analysis misses one crucial point: the methodology and legitimacy of DataSafe Systems and Ann NAME_2.

Given these considerations, I must think critically about what constitutes a phishing email and whether this message aligns with those characteristics, beyond its apparent legitimacy on the surface.

Upon closer inspection, the true test lies in verifying the identity of DataSafe Systems, Ann NAME_2, and ensuring that all communications and offers are genuine and not part of a larger phishing campaign. Without this verification, it's challenging to conclusively label the email as a phishing attempt based solely on its content.

But, considering typical phishing tactics often aim to create a false sense of security or urgency, and given that some phishing emails may market a fake service or use fake company representatives, the cautious approach would be to consider the possibility that this email, despite its polished appearance, could potentially be a phishing email if it's part of a deceptive campaign.

The deciding factor should be whether the recipient recognizes the company and has an existing relationship with them, or if there have been previous communications that established trust. Without such context or verification, caution is advised.

So, is this email a phishing email? While it presents itself in a professional manner and doesn't contain the typical red flags (like suspicious links or demands for sensitive info), the lack of verification about the sender's identity and the legitimacy of their offer raises questions. Therefore, I would lean towards caution, but strictly based on the information provided and typical phishing email characteristics, it seems less likely to be a classic phishing attempt. However, this conclusion must be tempered with the understanding that phishing emails can be sophisticated and may not always follow the usual patterns.

Given the information and focusing strictly on the email's content and form, the inclination might be to say it's not a phishing email in the traditional sense, but the emphasis should be on verifying the sender's and company's legitimacy to avoid potential risks.

But, to directly address the query with the necessary response format, considering the potential for deception in such emails and the importance of verifying authenticity, the answer would reflect the cautious approach necessary in today's digital landscape.

True, this email could potentially be a phishing email if it's part of a deceptive campaign aimed at tricking the recipient into a false sense of security or into revealing information. The reasons include the lack of direct verification of the sender's and company's legitimacy, the potential for creating a false narrative about cybersecurity services, and the importance of being cautious with unsolicited calls or communications, even if they seem professional or offer a service. The onus is on the recipient to verify the authenticity of such communications before engaging further.
<|end_internal_thoughts|>

<|start_external_thoughts|>
True. This email could potentially be a phishing email for several reasons:
1. **Lack of Verification**: There's no direct way to verify the legitimacy of DataSafe Systems or Ann NAME_2 without external research or prior communication.
2. **Potential Misdirection**: The email discusses cybersecurity, a legitimate concern, but could be a tactic to create a false sense of security or urgency.
3. **Request for Engagement**: While it invites a conversation, which could be genuine, it also serves as a means to potentially trick the recipient into disclosing information over the phone.
4. **Absence of Red Flags Doesn't Confirm Legitimacy**: The lack of suspicious links, attachments, or demands doesn't conclusively prove the email's legitimacy, as phishing tactics are evolving.

To safely engage with such communications, it's crucial to verify the sender's identity and the company's existence independently before responding or sharing any information.
<|end_external_thoughts|>
\end{Verbatim}








\end{document}